%% file: main.tex
\documentclass[10pt,twocolumn,letterpaper]{article}

\usepackage{iccv}
\usepackage{times}
\usepackage{epsfig}
\usepackage{graphicx}
\usepackage{amsmath}
\usepackage{amssymb}
\usepackage{booktabs}
\usepackage{multirow}
\usepackage{makecell}

\usepackage[pagebackref=true,breaklinks=true,letterpaper=true,colorlinks,bookmarks=false,citecolor=blue,linkcolor=blue]{hyperref}

\iccvfinalcopy 

\def\iccvPaperID{1061} 

\def\etal{\emph{et~al.}}		     
\def\eg{\emph{e.g.,~}}               
\def\ie{\emph{i.e.,~}}               

\ificcvfinal\fi

\begin{document}

\title{Understanding Synonymous Referring Expressions via Contrastive Features}

\author{Yi-Wen Chen$^1$, Yi-Hsuan Tsai$^2$, Ming-Hsuan Yang$^{1,3}$ \\
University of California, Merced$^1$, NEC Labs America$^2$, Google Research$^3$}

\maketitle
\ificcvfinal\thispagestyle{empty}\fi

\input{0_abstract}

\input{1_intro}

\input{2_related}

\input{3_method}

\input{4_result}

\input{5_conclusion}

{\small
\bibliographystyle{ieee_fullname}
\bibliography{mybib}
}

\clearpage

\appendix
\input{6_appendix}
\end{document}

%% file: 0_abstract.tex
\begin{abstract}
Referring expression comprehension aims to localize objects identified by natural language descriptions.
This is a challenging task as it requires understanding of both visual and language domains.
%
%
One nature is that each object can be described by synonymous sentences with paraphrases, and such varieties in languages have critical impact on learning a comprehension model.
While prior work usually treats each sentence and attends it to an object separately,
%
we focus on learning a referring expression comprehension model that considers the property in synonymous sentences.
%
%
To this end, we develop an end-to-end trainable framework to learn contrastive features on the image and object instance levels, where features extracted from synonymous sentences to describe the same object should be closer to each other after mapping to the visual domain.
%
%
%
We conduct extensive experiments to evaluate the proposed algorithm on several benchmark datasets, and demonstrate that our method performs favorably against the state-of-the-art approaches.
Furthermore, since the varieties in expressions become larger across datasets when they describe objects in different ways, we present the cross-dataset and transfer learning settings to validate the ability of our learned transferable features.

\end{abstract}

%% file: 1_intro.tex
\vspace{-3mm}
\section{Introduction}
\vspace{-1mm}
Referring expression comprehension is a task to localize a particular object within an image guided by a natural language description, \eg ``the man holding a remote standing next to a woman'' or ``the blue car''.
Since referring expressions are widely used in our daily conversations, the ability to understand such expressions provides an intuitive way for humans to interact with intelligent agents.
One challenge of this task is to jointly comprehend the knowledge from both visual and language domains, where there are multiple ways (\ie synonymous sentences) to describe and paraphrase the same object. 
For instance, we can refer to an object by its attribute, location, or interaction with other objects.  
Referring expressions also vary in lengths and synonyms. 
As such, the varieties of sentences that describe the same object cause gaps in the language domain, and affect the model training process.
%
%

\begin{figure}[t]
	\centering
	\includegraphics[width=\linewidth]{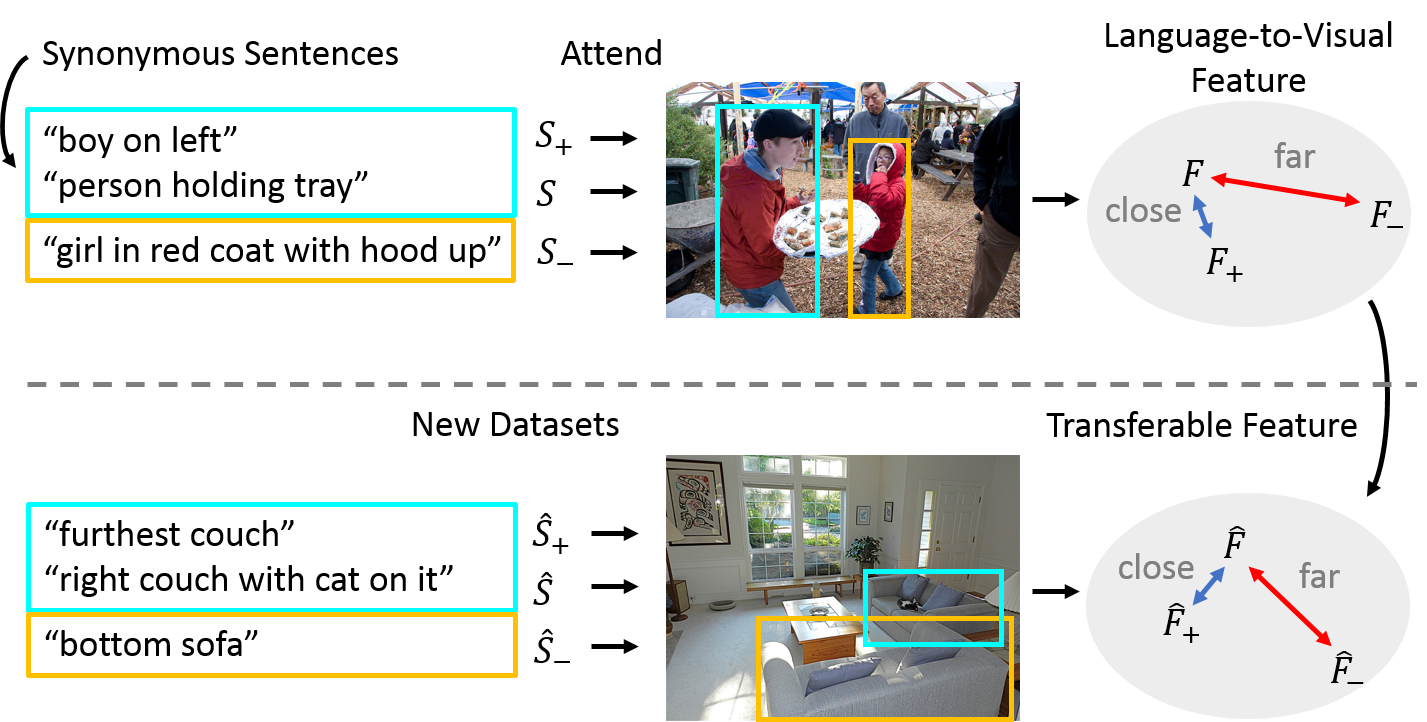}
	\caption{\textbf{Overview of the proposed algorithm.} 
	%
	%
	An object can be described in different ways, \eg by its attribute, location, or interaction with other objects. For a referring expression $S$, there are positive expressions $S_+$ describing the same object and negative ones $S_-$ for another object.
	While prior work considers each expression way separately, our method encourages features of synonymous sentences for the same object to attend nearby in the language-to-visual embedding space ($F$ and $F_+$) but far away from negative ones ($F_-$). Thus, the proposed framework can transfer learned features to unseen data.}
	\label{fig:overview}
	\vspace{-4mm}
\end{figure}

In this work, we take this language property, \textit{synonymous sentences}, into consideration during the training process.
This is different from existing referring expression comprehension methods \cite{Yu_CVPR_2018,Zhang_CVPR_2018,Liu_CVPR_2019,Yang_ICCV_2019} that consider all the sentences as individual instances.
Here, our main idea is to learn contrastive features when mapping the language features to the visual domain.
%
That is, while the same object can be described by different synonymous sentences, these language features should be close to each other after mapping to the visual domain.
On the other hand, for other expressions that do not describe that object, our model should also map them further away from that object (see Figure \ref{fig:overview} for an illustration).

To exploit how synonymous sentences are utilized to help model training as described above, we integrate feature learning techniques, \eg contrastive learning \cite{Hadsell_cvpr_06,info_nce,moco}, into our framework.
Then, the requirement of (multiple) positive/negative samples in contrastive learning can be satisfied by the notion of synonymous sentences.
However, it is not trivial to determine where we learn contrastive features in the model, in which we find that using language-to-visual features is beneficial to optimizing both the image and language modules.
To this end, we design an end-to-end learnable framework that enables feature learning on two different levels with language-to-visual features, \ie image and object instance levels, which are responsible for global context and relationships between object instances, respectively.
Moreover, since there are large varieties of negative samples (\ie any languages describing different objects can be negatives), we explore the option of mining negative samples to further facilitate the learning process.

In our framework, one benefit of learning contrastive features from synonymous sentences is to equip the model with the ability to contrast different language meanings. This ability is important when transferring the model to other datasets, as each domain may contain different varieties of sentences to describe objects.
To understand whether the learned features are effectively transferred to a new domain, we show that our model performs better in the cross-dataset setting (testing on the unseen dataset), as well as in the transfer learning setting that fine-tunes our pre-trained model on the target dataset.
Note that, although a similar concept of synonymous sentences is also adopted in  \cite{Wang_CVPR_2016} for retrieval tasks, it has not been exploited in referring expression comprehension for feature learning and transfer learning.
Specifically, \cite{Wang_CVPR_2016} uses off-the-shelf feature extractors and does not consider feature learning using multiple positive/negative samples on both image and instance levels like our framework.

We conduct extensive experiments on referring expression comprehension benchmarks to demonstrate the merits of learning contrastive features from synonymous sentences.
First, we use the RefCOCO benchmarks, including RefCOCO \cite{Yu_ECCV_2016}, RefCOCO+ \cite{Yu_ECCV_2016}, and RefCOCOg \cite{Mao_CVPR_2016,Nagaraja_ECCV_2016}, to perform baseline studies with comparisons to state-of-the-art methods.
Second, we focus on cross-dataset and transfer learning settings using the ReferItGame~\cite{Kazemzadeh_EMNLP_2014} and Ref-Reasoning~\cite{Yang_CVPR_2020} datasets to validate the ability of transferable features learned on the RefCOCO benchmarks.

The main contributions of this work are summarized as follows:
1) We propose a unified and end-to-end learnable framework for referring expression comprehension by considering various synonymous sentences to improve the training procedure.
2) We integrate feature learning techniques into our framework with a well-designed sampling strategy that learns contrastive features on both the image and instance levels.
%
%
3) We demonstrate that our model is able to effectively transfer learned representations in both cross-dataset and transfer learning settings.

%% file: 2_related.tex
\vspace{-1mm}
\section{Related Work}
\vspace{-1mm}

{\flushleft {\bf Referring Expression Comprehension.}}
%
%
The task of referring expression comprehension is typically considered as determining an object among several object proposals, given the referring expression.
To this end, several methods adopt two-stage frameworks to first generate object proposals with a pre-trained object detection network, and then rank the proposals according to the expression.
For example, CNN-LSTM models have been used to generate captions based on the image and proposals \cite{Hu_CVPR_2016_2,Luo_CVPR_2017,Mao_CVPR_2016}, and the one with the maximum posterior probability for generating the query expression is selected.
Other approaches~\cite{Wang_CVPR_2016,Rohrbach_ECCV_2016} embed proposals and the query sentence into a common feature space, and choose the object with the minimum distance to the expression.
In addition, several strategies are adopted to improve the performance, \eg analyzing the relationship between an object and its context~\cite{Nagaraja_ECCV_2016,Yu_ECCV_2016,Zhang_CVPR_2018}, or exploring the attributes~\cite{Liu_ICCV_2017_2} to distinguish similar objects.
To jointly consider multiple factors, MAttNet \cite{Yu_CVPR_2018} learns a modular network by considering three components, \ie subject appearance, location and relationship to other objects.
While these two-stage frameworks achieve promising results, the computational cost is significantly high due to extensive post-processing steps and individual models.
Furthermore, the model performance is largely limited by the pre-trained object detection network.
%

Some recent approaches adopt one-stage frameworks to tackle referring expression object segmentation \cite{Chen_BMVC_2019}, zero-shot grounding \cite{Sadhu_ICCV_2019}, and visual grounding \cite{Yang_ICCV_2019_2,Yang_ECCV_2020,Yang_ECCV_2020_2}, where the language features are fused with the object detector.
In these methods, the models are end-to-end trainable and more computationally efficient.
Compared to the methods mentioned above that consider each individual dataset separately, we aim to learn contrastive features by considering synonymous sentences during the training process.
In this work, we adopt a one-stage framework in which the representations in both the language and visual domains are learned jointly.

\begin{figure*}[tp]
	\centering
	\includegraphics[width=0.9\linewidth]{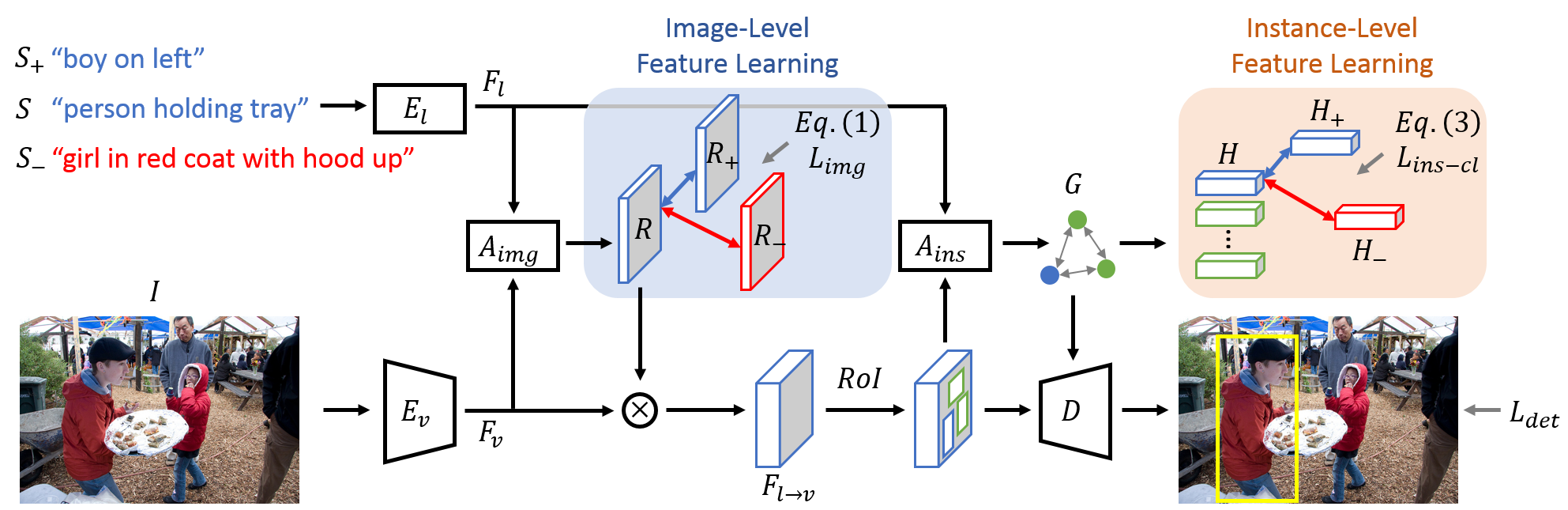}
	\caption{\textbf{Pipeline of the proposed framework.} 
	The features of the input image $I$ and referring expression $S$ (with its synonymous sentence $S_+$ and a negative expression $S_-$) are first extracted by a visual encoder $E_v$ and language encoder $E_l$, respectively. Then we adopt two attention modules $A_{img}$ and $A_{ins}$ for attending language features to the visual domain on the image and instance levels, where we apply our feature learning losses $L_{img}$ and $L_{ins-cl}$ to contrast positive/negative pairs, \ie $\{R_+, R_-\}$ and $\{H_+, H_-\}$ on two levels, respectively. On the instance level, a graph convolutional network $G$ is employed to model the relationships between object proposals. Finally, we generate the object bounding box with a detection head $D$.}
	\label{fig:framework}
	\vspace{-4mm}
\end{figure*}

\vspace{-1mm}
{\flushleft {\bf Feature Learning.}}
Feature learning aims to represent data in the embedding space, where similar data points are close to each other, and dissimilar ones are far apart, based on pairwise \cite{n_pair}, triplet \cite{triplet} or contrastive relationships \cite{Hadsell_cvpr_06,info_nce,moco}.
This representation model is then used for downstream tasks such as classification and detection.
%
%
%
These loss functions are computed on anchor, positive and negative samples, where the anchor-positive distance is minimized, and the anchor-negative distance is maximized.
While the triplet loss \cite{triplet} uses one positive and one negative sample per anchor, contrastive loss \cite{Hadsell_cvpr_06,info_nce,moco} includes multiple positive and negative samples for each anchor, which makes the learning process more efficient.
%
%
%

For natural language processing tasks, recent studies \cite{Peters_NAACL_2018,Devlin_arxiv_2018} based on the transformer \cite{Vaswani_NIPS_2017} have shown success in transfer learning.
Built upon the transformer-based BERT \cite{Devlin_arxiv_2018} model, learning representation for vision and language tasks by large-scale pre-training methods recently attracts much attention. 
These methods \cite{Lu_NIPS_2019,Lu_CVPR_2020,Zhou_AAAI_2020,Li_ECCV_2020,Chen_ECCV_2020,Gan_NIPS_2020} learn generic representations from a large amount of image-text pairs in a self-supervised manner, and fine-tune the model for the downstream vision and language tasks.
Recently, ViLBERT \cite{Lu_NIPS_2019} and its multi-tasking version \cite{Lu_CVPR_2020} use two parallel BERT-style models to extract features on image regions and text segments, and connect the two streams with co-attentional transformer layers.
%
%
Moreover, the OSCAR~\cite{Li_ECCV_2020} method uses object tags as anchor points to align the vision and language modalities in a shared semantic space.
While these approaches aim to learn generic representations for vision and language tasks by training the models on large-scale datasets, we focus on the referring expression comprehension, and adopt the feature learning techniques to improve the performance by considering synonymous sentences.

In this work, with a similar spirit to feature learning, we integrate the contrastive loss into our model by considering synonymous sentences for referring expression comprehension.
While it is natural to use the concept of synonymous expressions in tasks such as image-text retrieval \cite{Wang_CVPR_2016} or text-based person retrieval \cite{Yamaguchi_ICCV_2017}, it has not been exploited in referring expression comprehension to improve feature learning and further for transfer learning.
%
Different from \cite{Wang_CVPR_2016,Yamaguchi_ICCV_2017} that apply hinge loss on triplets, we consider multiple positive and negative samples for each anchor, and perform contrastive learning on both image and instance levels.
Moreover, we adopt an end-to-end framework, where the visual and language features are jointly learned, which is different from \cite{Wang_CVPR_2016,Yamaguchi_ICCV_2017} that use off-the-shelf feature extractors.



%% file: 3_method.tex
\vspace{-1mm}
\section{Proposed Framework}
\vspace{-1mm}

In this work, we address the problem of referring expression comprehension via using the information in \textit{synonymous sentences} to learn contrastive features across sentences.
Given an input image $I$ and a referring expression $S = \{w_t\}_{t=1}^T$ consisting of $T$ words $w_t$, the task is to localize the object identified by the expression.
We design a framework composed of a visual encoder $E_v$ and a language encoder $E_l$ for feature extraction in the visual and language domains.
Since our goal is to learn contrastive features after mapping the language features to the visual domain, we utilize the attention modules $A_{img}$ and $A_{ins}$ for language-to-visual features and a graph convolutional network (GCN) $G$ for aggregating instance-level features, which will be detailed in the following sections.
The output of referring expression comprehension is predicted by a detection head $D$.
Figure~\ref{fig:framework} shows the pipeline of the proposed framework.

Our method learns contrastive features on two levels to account for both the global context and relationships between object instances.
%
To this end, we enforce that the language-to-visual features on either the image or instance level should be close to each other if the referring expressions are synonymous sentences, and vice versa.
Given the image-level attention map $R_{l\rightarrow v}$ obtained from $A_{img}$ and the instance-level attention feature $H_{l\rightarrow v}$ inferred from $A_{ins}$ followed by a GCN module $G$, we regularize $R_{l\rightarrow v}$ and $H_{l\rightarrow v}$ by leveraging feature learning techniques, guided by the notion of synonymous sentences.
As a result, our language-to-visual features contrast the attentions from the language domain to the visual one based on the sentence meanings, which facilitates the comprehension task.

\subsection{Image-Level Feature Learning}
\label{sec:img_fea}
\vspace{-1mm}

To comprehend the information from the input image and referring expression, features of the two inputs are first extracted by each individual encoder.
We then use an attention module $A_{img}$ that attends the $l$-dimensional language feature $F_l = E_l(S) \in \mathbb{R}^l$ to the $v$-dimensional visual feature $F_v = E_v(I) \in \mathbb{R}^{h \times w \times v}$, where $h$ and $w$ denote the spatial dimension. 
A response map $R_{l\rightarrow v}$ that contains the multimodal information is obtained accordingly, \ie $R_{l\rightarrow v} = A_{img}(F_l, F_v) \in \mathbb{R}^{h \times w}$.
The details of the encoders and attention module $A_{img}$ are presented later in Section~\ref{sec:imp}.
%
%

As there are numerous synonymous sentences to describe the same object, the language-to-visual features should attend to similar regions regardless of how to describe the object.
%
%
%
Intuitively, We can apply a triplet loss on the response map to encourage the samples with synonymous expressions describing the same object to be mapped closely in the embedding space. Otherwise, they should be mapped far away from each other.

Specifically, for each input image $I$ and a referring expression anchor $S$, we randomly sample a positive expression $S_{+}$ that describes the same object, and a negative expression $S_{-}$ identifying a different object within the same image.
The triplet loss is then computed on the response generated by attending the three expression samples to the image $I$:
\begin{equation}
L_{img} = \max (d(R, R_{+}) - d(R, R_{-}) + \alpha, 0),
\label{eq:img_tri}
\end{equation}
where $R$, $R_{+}$ and $R_{-}$ are the responses of the anchor, positive and negative samples, respectively.
In addition, $d$ is the L2 distance between two response maps and $\alpha$ is the margin. 
After this step, we combine $R_{l\rightarrow v}$ and $F_v$ via element-wise multiplication 
$\otimes$ to produce the attentive feature $F_{l\rightarrow v} = R_{l\rightarrow v} \otimes F_v \in \mathbb{R}^{h \times w \times v}$, which is then used as the input to the detection head $D$ (see Figure \ref{fig:framework}).
We note that the triplet loss is applied to the response map $R_{l\rightarrow v}$ rather than the feature $F_{l\rightarrow v}$, since $R_{l\rightarrow v}$ is easier to optimize with the much lower dimension.

\subsection{Instance-Level Feature Learning}
\vspace{-1mm}

In addition to applying the triplet loss on the image level for learning contrastive features, we also consider the features on the instance level to encourage the model to focus more on local cues.
However, through the RoI module that generates proposals, each proposal (instance) only contains the information within its bounding box and does not provide the local context information (\eg interactions with other objects) described in the referring expression.
%
%
To tackle this problem, we design a graph convolutional network (GCN)~\cite{Kipf_ICLR_2017} in a way similar to \cite{Yang_ICCV_2019} to model the relationships between proposals, and then use the features after GCN as the input to our instance-level feature learning module.
More details regarding the implementation are provided in the supplementary material.

\vspace{-1mm}
{\flushleft {\bf Contrastive Feature Learning.}}
As shown in Figure \ref{fig:framework}, similar to the image-level in Section \ref{sec:img_fea}, we first adopt an instance-level attention module $A_{ins}$ following \cite{Yu_CVPR_2018} to attend the language feature $F_l$ to the RoI proposals, followed by the GCN module $G$ to aggregate the proposal relationships.
As a result, we obtain the instance-level language-to-visual features $H_{l\rightarrow v} = G(A_{ins}(F_l, \text{RoI}(F_{l\rightarrow v})))$.

%

%
%
Next, we propose to regularize $H_{l\rightarrow v}$ guided by the concept of synonymous sentences, where the proposal features from referring expressions that describe the same object should be close to each other. Otherwise, they should be apart from each other.
To this end, one straightforward way to learn instance-level contrastive features is to apply the triplet loss similar to \eqref{eq:img_tri}:
\begin{equation}
L_{ins-tri} = \max (d(H, H_{+}) - d(H, H_{-}) + \alpha, 0),
\label{eq:ins_tri}
\end{equation}
where $H$, $H_{+}$, and $H_{-}$ represent instance-level features of the anchor, positive and negative expressions attending on the visual domain, respectively.
%
%
%
%
Note that each expression may generate different RoI locations. To use the same proposals across samples in the triplet, we select the proposal with the highest IoU score with respect to the ground truth bounding box.

%
%
\vspace{-1mm}
{\flushleft {\bf Contrastive Loss with Negative Mining.}}
Although the triplet loss in \eqref{eq:ins_tri} can be used to learn instance-level contrastive features, it is limited to sample one positive and negative at a time, which may not fully exploit multiple synonymous sentences.
%
Therefore, we leverage the contrastive loss \cite{Khosla_nips_20} with the property that can consider multiple positive/negative samples.
Intuitively, we can treat synonymous sentences as positives, but the space of negative samples is large and noisy as any two languages describing different objects are negatives.
Moreover, it has been studied that finding good negative samples is critical for learning contrastive features effectively \cite{Kalantidis_nips_20}.
%
%

To tackle this issue, we employ the following two strategies to mine useful negative samples:
1) From the perspective of the visual domain, we mine samples that describe the same object category but in different images, and then use the corresponding referring expressions as the negatives.
This encourages our model to contrast features that describe similar contents across images, as sentences referring to the same object category usually share common contexts.
2) Considering the language embedding features, we mine the top $N$ samples (\ie $N = 8$ in this work) that have closer language features to the anchor sample but in different images. 
This helps the model contrast samples that have a similar language structure.
Overall, our contrastive loss $L_{ins-cl}$ can be formulated as:
\begin{align}
- \text{log} \frac{\sum\limits_{H_{+} \in Q_+} e^{h(H)^\top h(H_{+}) / \tau}}{\sum\limits_{H_{+} \in Q_+} e^{h(H)^\top h(H_{+})/ \tau} + \sum\limits_{H_{-} \in Q_-} e^{h(H)^\top h(H_{-})/ \tau}},
\label{eq:ins_cl}
\end{align}
where $Q_+$ and $Q_-$ are the sets of positive and negative samples, and $\tau$ is the temperature parameter. 
In practice, we follow the SimCLR \cite{simclr} method and use $h(\cdot)$ as a linear layer that projects features $H$ to an embedding space where the contrastive loss is applied.
%
\vspace{-1mm}
{\flushleft {\bf Discussions.}}
Compared to the instance-level triplet loss in \eqref{eq:ins_tri}, using the contrastive loss in \eqref{eq:ins_cl} has a few merits. 
First, given an anchor sample, it can contrast with multiple positive and negative samples at the same time, which is much more efficient than sampling the triplets, as shown in \cite{Khosla_nips_20} and our ablation study presented later. 
%
Second, the projection head $h(\cdot)$ provides a learnable buffer before feeding features to compute the contrastive loss, which helps the model learn better representations.

In terms of the sampling strategy, different from the negative mining method in \cite{Chen_CVPR_2020} that generates negatives with the same object category, we also consider negatives with similar languages to the anchor language but containing other object categories.
This difference allows us to sample more negatives with similar language structures, which is crucial for our contrastive loss in language.


\subsection{Model Training and Implementation Details}
\vspace{-1mm}
\label{sec:imp}
In this section, we provide more details in training our framework and the design choices.
\vspace{-1mm}
{\flushleft {\bf Overall Objective.}}
The overall objective for the proposed algorithm consists of the aforementioned loss functions (\ref{eq:img_tri}) and (\ref{eq:ins_cl}) for learning contrastive features on the image and instance levels, and the detection loss $L_{det}$ as defined in the Mask R-CNN~\cite{He_ICCV_2017} following the MAttNet~\cite{Yu_CVPR_2018} method:
%
\begin{equation}
L_{all} = L_{det} + L_{img} + L_{ins-cl}.
\label{eq:loss_total}
\end{equation}
%
%
\vspace{-1mm}
{\flushleft {\bf Implementation Details.}}
%
%
For the visual encoder $E_v$ in our framework and the detection head $D$, we adopt the Mask R-CNN~\cite{He_ICCV_2017} as the backbone model, which is pre-trained on COCO training images, excluding those in validation and testing splits of RefCOCO, RefCOCO+ and RefCOCOg.
The ResNet-101~\cite{He_CVPR_2016} is used as the feature extractor, where the output of the final convolutional layer in the fourth block is the feature $F_v$ that serves as the input to the attention module $A_{img}$.
For the language encoder $E_l$, we use either the BERT~\cite{Devlin_arxiv_2018} or Bi-LSTM model.
%
%
In the image-level attention module $A_{img}$, we adopt dynamic filters similar to~\cite{Chen_BMVC_2019} to attend language features to the visual domain and generate $R_{l\rightarrow v}$.
%
%
On the instance level, features after GCN are duplicated to each spatial location and concatenated with the spatial features after RoI. Then, the concatenated features are fed to the detection head $D$ to generate final results.
During testing, the detected object with the largest score is considered as the prediction.
The margin $\alpha$ in the triplet loss (\ref{eq:img_tri}) and (\ref{eq:ins_tri}) is set to 1. In the contrastive loss (\ref{eq:ins_cl}), the temperature $\tau$ is set to 0.1, and the projection head $h(\cdot)$ is a 2-layer MLP, projecting the features to a 128-dimensional latent space.
We implement the proposed model in PyTorch with the SGD optimizer, and the entire model is trained end-to-end with 10 epochs. The initial learning rate is set to $10^{-4}$ and decreased to $10^{-5}$ after 3 epochs.
The source code and trained models are available at \url{https://github.com/wenz116/RefContrast}.






%% file: 4_result.tex
\vspace{-1mm}
\section{Experimental Results}
\vspace{-1mm}
We evaluate the proposed framework on three referring expression datasets, including RefCOCO~\cite{Yu_ECCV_2016}, RefCOCO+~\cite{Yu_ECCV_2016} and RefCOCOg~\cite{Mao_CVPR_2016,Nagaraja_ECCV_2016}.
The three datasets are collected on the MSCOCO~\cite{MSCOCO} images, but with different ways to generate referring expressions.
Extensive experiments are conducted in multiple settings.
We first compare the performance of the proposed algorithm with state-of-the-art methods and present the ablation study to show the improvement made by each component.
%
%
Then we evaluate the models on the unseen Ref-Reasoning~\cite{Yang_CVPR_2020} dataset to validate the effectiveness on unseen datasets.
Furthermore, we conduct experiments in the transfer learning setting, where the pre-trained models are fine-tuned on either the Ref-Reasoning~\cite{Yang_CVPR_2020} or ReferItGame~\cite{Kazemzadeh_EMNLP_2014} dataset.
More results and analysis are provided in the supplementary material.
%
\vspace{-1mm}
{\flushleft {\bf Intra- and Inter-Dataset Feature Learning.}}
Since synonymous sentences exist within one dataset and across datasets,
we consider both intra- and inter-dataset feature learning loss.
For each input image and referring expression anchor, we sample positive and negative expressions from the same dataset for the intra-dataset case, and expressions from different datasets as inter-dataset samples.
%
%
%
In our experiments, we use the intra-dataset loss for training on a single dataset, and both the intra- and inter-dataset losses for jointly training on the three datasets.

\begin{table*}[!t]
	\centering
	\footnotesize
	\caption{\textbf{Comparisons with state-of-the-art methods.}
	The models in the top and middle groups are trained on a single dataset, while those in the bottom group are jointly trained on the RefCOCO, RefCOCO+ and RefCOCOg* (our split) datasets, where same images are shared among three datasets but with more expressions than the top and middle groups.}
	\begin{tabular}{ccccccccccc}
		\toprule
		& & \multicolumn{3}{c}{RefCOCO} & \multicolumn{3}{c}{RefCOCO+} & \multicolumn{3}{c}{RefCOCOg} \\
		\cmidrule(lr){3-5} \cmidrule(lr){6-8} \cmidrule(lr){9-11}
		Method & Language Encoder & val & testA & testB & val & testA & testB & val* & val & test \\ \midrule
		Nagaraja~\etal~\cite{Nagaraja_ECCV_2016} & LSTM & 57.30 & 58.60 & 56.40 & -  & - & -  & -  & - & 49.50 \\
		Luo~\etal~\cite{Luo_CVPR_2017}       & LSTM & - & 67.94 & 55.18 & - & 57.05 & 43.33 & 49.07 & - & - \\
		SLR~\cite{Yu_CVPR_2017}              & LSTM & 69.48 & 73.71 & 64.96 & 55.71 & 60.74 & 48.80 & - & 60.21 & 59.63 \\
		Liu~\etal~\cite{Liu_ICCV_2017_2}     & LSTM & - & 72.08 & 57.29 & - & 57.97 & 46.20 & 52.35 & - & - \\
		PLAN~\cite{Zhuang_CVPR_2018}         & LSTM & - & 75.31 & 65.52 & - & 61.34 & 50.86 & 58.03 & - & - \\
		VC~\cite{Zhang_CVPR_2018}            & LSTM & - & 73.33 & 67.44 & - & 58.40 & 53.18 & 62.30 & - & - \\
		MAttNet~\cite{Yu_CVPR_2018}          & LSTM & 76.65 & 81.14 & 69.99 & 65.33 & 71.62 & 56.02 & - & 66.58 & 67.27 \\
		CM-Att-Erase~\cite{Liu_CVPR_2019}    & LSTM & 78.35 & \textbf{83.14} & 71.32 & \textbf{68.09} & 73.65 & 58.03 & - & 67.99 & 68.67 \\
		CAC~\cite{Chen_BMVC_2019}            & LSTM & 77.08 & 80.34 & 70.62 & 62.15 & 69.26 & 51.32 & 62.34 & 65.83 & 65.44 \\
		DGA~\cite{Yang_ICCV_2019}            & LSTM & - & 78.42 & 65.53 & - & 69.07 & 51.99 & - & - & 63.28 \\
		Darknet-LSTM~\cite{Yang_ICCV_2019_2} & LSTM & 73.66 & 75.78 & 71.32 & - & - & - & - & - & - \\
		RCCF~\cite{Liao_CVPR_2020}           & LSTM & - & 81.06 & 71.85 & - & 70.35 & 56.32 & - & - & 65.73 \\
		Ours (baseline)                      & LSTM & 76.63 & 79.59 & 69.93 & 63.67 & 71.26 & 55.04 & 61.71 & 66.24 & 65.85 \\
		Ours (full model)                    & LSTM & \textbf{79.09} & 82.86 & \textbf{72.64} & 66.81 & \textbf{74.23} & \textbf{58.08} & \textbf{64.80} & \textbf{69.14} & \textbf{68.86} \\ \midrule
		
		Darknet-BERT~\cite{Yang_ICCV_2019_2} & BERT & 72.05 & 74.81 & 67.59 & 55.72 & 60.37 & 48.54 & 48.14 & 59.03 & 58.70 \\
		ReSC-Large~\cite{Yang_ECCV_2020}     & BERT & 77.63 & 80.45 & 72.30 & 63.59 & 68.36 & 56.81 & 63.12 & 67.30 & 67.20 \\
		Ours (baseline)                      & BERT & 77.02 & 80.25 & 70.53 & 64.31 & 71.83 & 55.20 & 62.27 & 66.48 & 66.34 \\
		Ours (full model)                    & BERT & \textbf{79.63} & \textbf{83.32} & \textbf{73.27} & \textbf{67.49} & \textbf{74.85} & \textbf{58.42} & \textbf{65.23} & \textbf{69.67} & \textbf{69.51} \\ \midrule
		Ours (baseline-all)                  & BERT & 75.17 & 79.53 & 68.72 & 63.42 & 70.55 & 53.38 & - & - & - \\
		Ours (full model-all)                & BERT & \textbf{82.42} & \textbf{85.77} & \textbf{75.29} & \textbf{70.64} & \textbf{78.12} & \textbf{61.49} & - & - & - \\

		\bottomrule
		\label{tab:comparison}
	\end{tabular}
	\vspace{-5mm}
\end{table*}

\begin{figure*}[!t]
    \footnotesize
	\centering
	\begin{tabular}
		{ @{\hspace{0mm}}c@{\hspace{1mm}} @{\hspace{0mm}}c@{\hspace{1mm}} @{\hspace{0mm}}c@{\hspace{1mm}} @{\hspace{0mm}}c@{\hspace{1mm}} @{\hspace{0mm}}c@{\hspace{1mm}} @{\hspace{0mm}}}
		\makecell{``man in maroon shirt\\wearing sunglasses''}  & \makecell{``woman in black\\not with child''} & \makecell{``orange cat behind''} & \makecell{``bike next to the girl rider''} & \makecell{``container opposite\\of the grapes''} \\
		\includegraphics[height=0.135\linewidth]{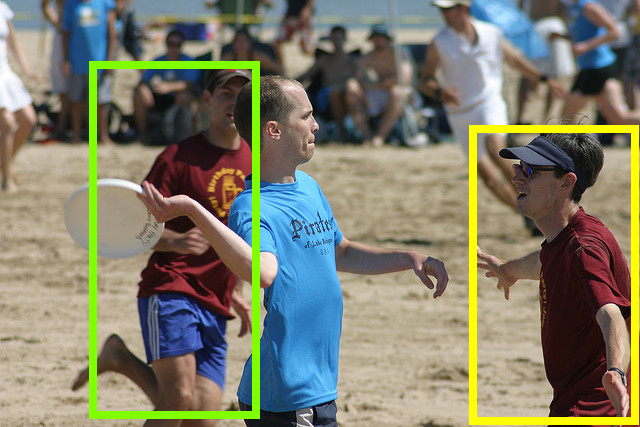} &	
		\includegraphics[height=0.135\linewidth]{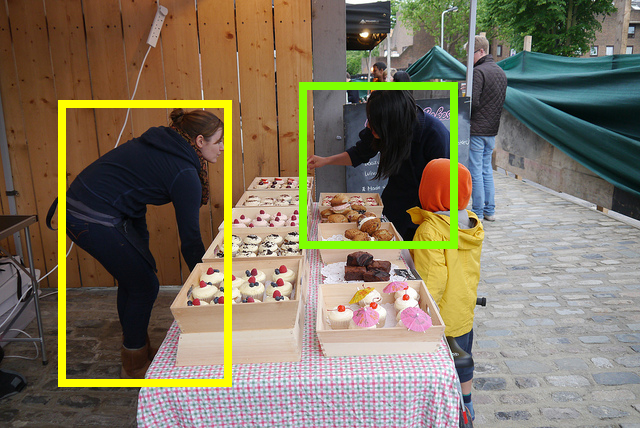} &
		\includegraphics[height=0.135\linewidth]{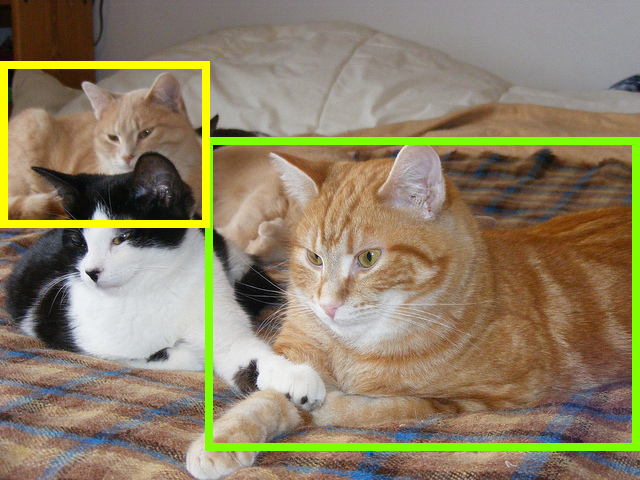} &
		\includegraphics[height=0.135\linewidth]{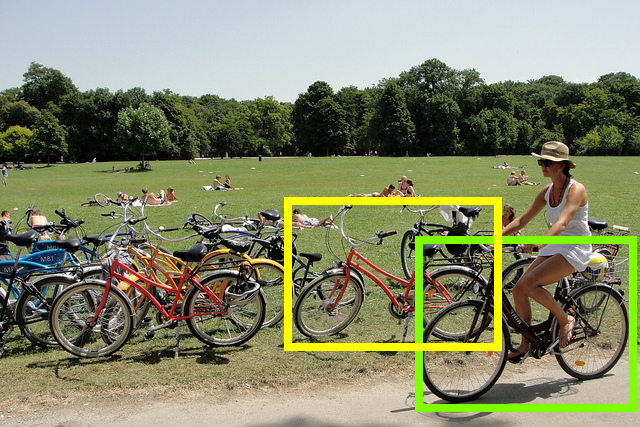} &
		\includegraphics[height=0.135\linewidth]{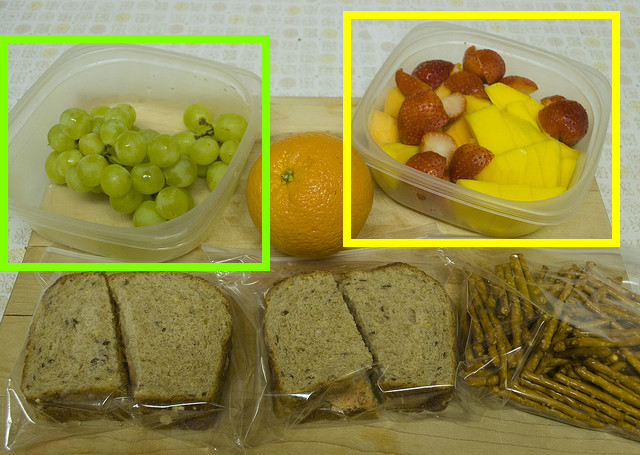} \\
	\end{tabular}
	\caption{\textbf{Sample results of jointly training on the RefCOCO, RefCOCO+ and RefCOCOg* datasets.} The green and yellow boxes represent the results of the baseline and our full model, respectively.}
	\label{fig:visualization}
	\vspace{-4mm}
\end{figure*}
%

\subsection{Datasets and Evaluation Metric}
\vspace{-1mm}

\textbf{RefCOCO} contains 19,994 images with 142,209 referring expressions for 50,000 objects, while \textbf{RefCOCO+} is composed of 141,564 expressions for 49,856 objects in 19,992 images.
Restrictions are not placed on generating expressions for RefCOCO, but put on RefCOCO+ by forbidding the location information, making it focus more on the appearance of the target object and its interaction with others.
The two testing splits testA and testB are generated respectively on images containing multiple people and images containing multiple objects of other categories.
We follow the split of the training, validation and testing images in \cite{Yu_ECCV_2016}, and there is no overlap across the three sets.
%

%
The \textbf{RefCOCOg} dataset consists of 85,474 referring expressions for 54,822 objects in 26,711 images with longer expressions.
There are two splits constructed in different ways.
The first split~\cite{Mao_CVPR_2016} randomly partitions objects into training and validation sets. Therefore, the same image could appear in both sets.
The validation set is denoted as ``val*'' in this paper.
%
The second partition~\cite{Nagaraja_ECCV_2016} randomly splits images into training, validation and testing sets, where we denote the validation and testing splits as ``val'' and ``test'', respectively.
In our experiments, when jointly training on three datasets, to avoid overlaps between the training, validation and testing images, we create another split for RefCOCOg, where each set contains the images present in the corresponding set of RefCOCO and RefCOCO+. We denote this split as ``RefCOCOg*''.

The \textbf{Ref-Reasoning} dataset contains 83,989 images from GQA~\cite{Hudson_CVPR_2019} with 791,956 referring expressions automatically generated based on the scene graphs.
%
The \textbf{ReferItGame} dataset is collected from ImageCLEF~\cite{Escalante_CVIU_2010} and consists of 130,525 expressions, referring to 96,654 objects in 19,894 images.
To evaluate the detection performance, the predicted bounding box is considered correct if the intersection-over-union (IoU) of the prediction and the ground truth bounding box is above 0.5.

\begin{table*}[t]
	\centering
 	\footnotesize
	\renewcommand{\arraystretch}{1.1}
	\caption{\textbf{Ablation study of jointly training on three datasets.} The top and middle groups demonstrate the effectiveness of the proposed feature learning techniques in different levels, and the superiority of contrastive loss over triplet loss in the instance level. The bottom group shows the influence of the negative mining and GCN in our model.}
	\begin{tabular}{cccccccccccc}
		\toprule
		& & & \multicolumn{3}{c}{RefCOCO} & \multicolumn{3}{c}{RefCOCO+} & \multicolumn{3}{c}{RefCOCOg*} \\ 
		\cmidrule(lr){4-6} \cmidrule(lr){7-9} \cmidrule(lr){10-12}
		Method & Image-level & Instance-level & val & testA & testB & val & testA & testB & val & testA & testB \\ 
		\midrule
		Baseline & & & 75.17 & 79.53 & 68.72 & 63.42 & 70.55 & 53.38 & 60.62 & 64.57 & 53.81 \\
		
		w/o Instance-level & \checkmark &  & 77.82 & 81.34 & 70.94 & 66.25 & 73.18 & 55.91 & 63.84 & 67.12 & 57.46 \\
		
		w/o Image-level &  & Triplet Loss \eqref{eq:ins_tri} & 78.49 & 81.92 & 71.35 & 67.04 & 74.80 & 57.32 & 64.58 & 67.79 & 58.31 \\
		
		w/o Image-level &  & Contrastive Loss \eqref{eq:ins_cl} & 79.33 & 83.28 & 72.64 & 67.81 & 75.77 & 58.63 & 65.73 & 68.54 & 58.89 \\
		
		\midrule
		
		Final w/ $L_{ins-tri}$ & \checkmark & Triplet Loss \eqref{eq:ins_tri} & 80.61 & 84.19 & 73.71 & 69.28 & 76.72 & 59.80 & 68.26 & 69.81 & 61.34 \\
		
		Final w/ $L_{ins-cl}$ & \checkmark & Contrastive Loss \eqref{eq:ins_cl} & \textbf{82.42} & \textbf{85.77} & \textbf{75.29} & \textbf{70.64} & \textbf{78.12} & \textbf{61.49} & \textbf{69.92} & \textbf{71.75} & \textbf{62.68} \\
		
		\midrule
		
		Final w/o Neg. Mining & \checkmark & Contrastive Loss \eqref{eq:ins_cl} & 82.05 & 85.52 & 74.89 & 70.38 & 77.82 & 61.17 & 69.46 & 71.32 & 62.23 \\
		
		Final w/o GCN & \checkmark & Contrastive Loss \eqref{eq:ins_cl} & 81.40 & 84.26 & 74.52 & 69.72 & 77.18 & 60.57 & 69.13 & 70.71 & 61.67 \\
		
		
		\bottomrule
		\label{tab:ablation}
	\end{tabular}
	\vspace{-6mm}
\end{table*}

\begin{table}[!t]
	\centering
	\footnotesize
	\setlength\tabcolsep{4.5pt}
	\caption{\textbf{Evaluation on the Ref-Reasoning dataset with models trained on different datasets.} The models are trained on the Ref-Reasoning (top group), RefCOCOg (middle group), or RefCOCO, RefCOCO+ and RefCOCOg* datasets (bottom group).}
	\begin{tabular}{ccccc}
		\toprule
		& & \multicolumn{3}{c}{Number of Objects} \\ \cmidrule{3-5} 
		Method & Training Dataset & one & two & three \\ \midrule
		
		Ours (supervised) & Ref-Reasoning & 76.43 & 57.37 & 50.79 \\
		
		\midrule
		MAttNet~\cite{Yu_CVPR_2018}       & \multirow{4}{*}{RefCOCOg}     & 49.81 & 32.17 & 25.83 \\
		CM-Att-Erase~\cite{Liu_CVPR_2019} &                               & 50.34 & 32.42 & 26.02 \\
		Ours (baseline)   &                                               & 50.26 & 31.41 & 26.16 \\
		Ours (full model) &                                               & \textbf{53.49} & \textbf{33.61} & \textbf{28.09} \\ \midrule
		Ours (baseline)   & \multirow{2}{*}{\makecell{RefCOCO, RefCOCO+,\\RefCOCOg*}} & 54.56 & 33.11 & 28.46 \\
		Ours (full model) &                                               & \textbf{57.96} & \textbf{36.82} & \textbf{32.61} \\
		\bottomrule
		\label{tab:unseen}
	\end{tabular}
	\vspace{-6mm}
\end{table}

\subsection{Evaluation on Seen Datasets}
\vspace{-1mm}

Table~\ref{tab:comparison} shows the results of the proposed algorithm against state-of-the-art methods~\cite{Nagaraja_ECCV_2016,Luo_CVPR_2017,Yu_CVPR_2017,Liu_ICCV_2017_2,Zhuang_CVPR_2018,Zhang_CVPR_2018,Yu_CVPR_2018,Liu_CVPR_2019,Chen_BMVC_2019,Yang_ICCV_2019,Yang_ICCV_2019_2,Liao_CVPR_2020,Yang_ECCV_2020}.
All the compared approaches except for the recent methods \cite{Chen_BMVC_2019,Yang_ICCV_2019_2,Liao_CVPR_2020,Yang_ECCV_2020} adopt two-stage frameworks, where the prediction is chosen from a set of proposals.
Therefore, their models are not end-to-end trainable, while our one-stage framework is able to learn better feature representations by end-to-end training.
Moreover, all of these methods train on each dataset separately and do not consider the varieties of synonymous sentences within/across datasets.

The models in the top and middle groups of Table~\ref{tab:comparison} are trained and evaluated on the same single dataset. We separate the methods using different language encoders for fair comparisons.
The results show that the full model trained with our loss functions consistently improves the baseline model without considering synonymous sentences.
%
In addition, our method with either LSTM or BERT as the language encoder performs favorably against other approaches on most dataset splits.
%
Note that, the runtime of our unified framework (0.325 seconds per frame) is much faster than the two-stage  MAttNet~\cite{Yu_CVPR_2018} and CM-Att-Erase~\cite{Liu_CVPR_2019} methods (0.671 and 0.734 seconds, respectively).

%

The bottom group of Table~\ref{tab:comparison} shows the results of our models jointly trained on the RefCOCO, RefCOCO+ and RefCOCOg* (our split) datasets.
%
Since the three datasets share the same images but contain expressions of very different properties, directly training on all of them as the baseline would cause training difficulties in a single model due to the large varieties in language.
%
In contrast, by applying the proposed loss terms, the performance improves from the baseline, and the gains are larger than those in the single dataset setting.
We also note that all the training images are the same, but the varieties of synonymous sentences are different across these two settings.

%

%

\subsection{Ablation Study}
\vspace{-1mm}

We present the ablation study results in Table~\ref{tab:ablation} to show the effect of each component in the proposed framework. The models are jointly trained on the RefCOCO, RefCOCO+ and RefCOCOg* datasets.
In the top group of Table~\ref{tab:ablation}, we first demonstrate that applying either the image-level or instance-level loss improves the performance from the baseline model that does not consider the property of synonymous sentences.
In the middle group of the table, the results are further improved in the full models with the loss on both levels, and the one with contrastive loss in the instance level achieves better performance than the one using triplet loss.
%
In the bottom group, we provide the detailed ablation study in our model. When removing negative mining or GCN from the model, the performance is slightly reduced compared to the full model in the middle group.
%
This shows that our proposed feature learning techniques play the main role in performance improvement, while the proposed strategies regarding negative mining and GCN also help model training.
%
Furthermore, we present qualitative results in Figure~\ref{fig:visualization}, which show that our full model better distinguishes between similar objects and understands relationships across objects.
%









%
%
%
%
%
%



\subsection{Evaluation on the Unseen Dataset}
\vspace{-1mm}

To demonstrate that the proposed framework is able to transfer learned features to other unseen datasets, we use our trained models to evaluate on the Ref-Reasoning~\cite{Yang_CVPR_2020} dataset, which contains completely different images and expressions from our training datasets.
The results are shown in Table~\ref{tab:unseen} and the performance of our model trained on the Ref-Reasoning dataset using the fully-supervised setting is provided as a reference.
%
We first train our models on the RefCOCOg dataset~\cite{Nagaraja_ECCV_2016}.
%
When applying the intra-dataset loss in the full model, the performance is improved from the baseline and better than the two-stage MAttNet~\cite{Yu_CVPR_2018} and CM-Att-Erase~\cite{Liu_CVPR_2019} approaches, where we evaluate using their official pre-trained models.
Then we jointly train our models on RefCOCO, RefCOCO+ and RefCOCOg* (our split) datasets.
When more datasets are used to train our full model, the performance gains over the baseline increase, which demonstrates the effectiveness of our feature learning using synonymous sentences.
%

\begin{table}[!t]
	\centering
	\footnotesize
	\setlength\tabcolsep{4.5pt}
	\caption{\textbf{Transfer learning on the Ref-Reasoning dataset with different settings of pre-training (Pre) and fine-tuning (FT).} The models in the top group are directly trained on Ref-Reasoning, while those in the bottom group are pre-trained on RefCOCO, RefCOCO+ and RefCOCOg*, and fine-tuned on Ref-Reasoning. Note that * indicates that SGMN~\cite{Yang_CVPR_2020} uses ground truth proposals to generate final outputs, which is served as a reference here.}
	\begin{tabular}{ccccccc}
		\toprule
		& Pre-training & \multicolumn{2}{c}{Our loss in} & \multicolumn{3}{c}{Number of Objects} \\ \cmidrule(lr){3-4} \cmidrule(lr){5-7} 
		Method & Dataset & Pre & FT & one & two & three \\ \midrule
		SGMN~\cite{Yang_CVPR_2020}* & & & & 80.17 & 62.24 & 56.24 \\
		Ours (baseline)            &  & & & 76.43 & 57.37 & 50.79 \\ \midrule
		Ours (baseline)  & \multirow{4}{*}{\makecell{RefCOCO,\\RefCOCO+,\\RefCOCOg*}} & & & 76.54 & 57.43 & 50.81 \\
		Ours    & & & \checkmark & 78.72 & 59.14 & 52.03 \\
		Ours    & & \checkmark & & 79.16 & 59.47 & 52.39 \\
		Ours (full model) & & \checkmark & \checkmark & \textbf{81.94} & \textbf{62.73} & \textbf{55.86} \\
		\bottomrule
		\label{tab:transfer_ref_reasoning}
	\end{tabular}
	\vspace{-4mm}
\end{table}

\begin{table}[!t]
	\centering
	\footnotesize
	\caption{\textbf{Transfer learning on the ReferItGame dataset with different settings of pre-training (Pre) and fine-tuning (FT).} The models in the top group are directly trained on ReferItGame, while those in the bottom group are pre-trained on RefCOCO, RefCOCO+ and RefCOCOg*, and fine-tuned on ReferItGame.}
	\begin{tabular}{ccccc}
		\toprule
		& Pre-training & \multicolumn{2}{c}{Our loss in} & Split \\ \cmidrule(lr){3-4} \cmidrule(lr){5-5}
		Method & Dataset & Pre & FT & test \\ \midrule
		ZSGNet~\cite{Sadhu_ICCV_2019}        & & & & 58.63 \\
		Darknet-LSTM~\cite{Yang_ICCV_2019_2} & & & & 58.76 \\
		Darknet-BERT~\cite{Yang_ICCV_2019_2} & & & & 59.30 \\
		Ours (baseline)   & & & & 57.04 \\ \midrule
		Ours (baseline)   & \multirow{4}{*}{\makecell{RefCOCO,\\RefCOCO+,\\RefCOCOg*}} & & & 57.13 \\
		Ours       & & & \checkmark & 58.54 \\
		Ours       & & \checkmark & & 58.89 \\
		Ours (full model) & & \checkmark & \checkmark & \textbf{61.71} \\
		\bottomrule
		\label{tab:transfer_referitgame}
	\end{tabular}
	\vspace{-6mm}
\end{table}

\subsection{Transfer Learning on Unseen Datasets}
\vspace{-1mm}

To validate the feature learning ability of the proposed method, we conduct experiments on the transfer learning setting, where the models are pre-trained on the RefCOCO, RefCOCO+ and RefCOCOg* datasets, and fine-tuned on the Ref-Reasoning~\cite{Yang_CVPR_2020} or ReferItGame~\cite{Kazemzadeh_EMNLP_2014} dataset.
We also note that the SGMN~\cite{Yang_CVPR_2020} model in Table~\ref{tab:transfer_ref_reasoning} uses ground truth proposals from the dataset, which is served as a reference, but is not a direct comparison with our method that predicts the locations without accessing ground truth bounding boxes.

In Table~\ref{tab:transfer_ref_reasoning}, we first observe that two models of ``Ours (baseline)'', with or without the pre-training stage, perform very similarly to each other on Ref-Reasoning~\cite{Yang_CVPR_2020}.
%
This shows that it is not trivial to learn transferable features by simply pre-training on existing datasets.
However, by introducing our feature learning schemes, either during pre-training (Pre) or fine-tuning (FT), our models achieve better performance.
When using our method in both pre-training and fine-tuning stages, the performance is further improved.
%
%

Similar comparisons can be observed in Table~\ref{tab:transfer_referitgame}, where the models are fine-tuned on the ReferItGame \cite{Kazemzadeh_EMNLP_2014} dataset. 
In the bottom group, we demonstrate that the feature learning technique improves the performance consistently by using the proposed loss terms.
Compared to results in the top group, despite that our baseline model does not perform better than ZSGNet~\cite{Sadhu_ICCV_2019} and Darknet-LSTM~\cite{Yang_ICCV_2019_2}, we show significant improvement by using our proposed loss functions to achieve better performance. This shows the ability of our method for transferring learned representations to other datasets.
%

\begin{table}[!t]
	\centering
	\footnotesize
	\caption{\textbf{Similarity between instance-level features of synonymous sentences.} The models are jointly trained on the RefCOCO, RefCOCO+ and RefCOCOg* datasets.}
	\begin{tabular}{cccc}
		\toprule
		Method & RefCOCO & RefCOCO+ & RefCOCOg* \\ \midrule
		Ours (baseline)   & 0.884 & 0.873 & 0.851 \\
		Ours (full model) & 0.926 & 0.911 & 0.885 \\

		\bottomrule
		\label{tab:dot_joint}
	\end{tabular}
	\vspace{-4mm}
\end{table}

\begin{table}[!t]
	\centering
	\footnotesize
	\caption{\textbf{Similarity between instance-level features of synonymous sentences.} All models are jointly trained on the RefCOCO, RefCOCO+ and RefCOCOg* datasets. Only models in the bottom group are fine-tuned on Ref-Reasoning or ReferItGame.}
	\begin{tabular}{cccc}
		\toprule
		Method & Fine-tune & Ref-Reasoning & ReferItGame \\ \midrule
		Ours (baseline)   & & 0.762 & 0.787 \\
		Ours (full model) & & 0.781 & 0.804 \\ \midrule
		Ours (baseline)   & \checkmark & 0.843 & 0.856 \\
		Ours (full model) & \checkmark & 0.881 & 0.892 \\

		\bottomrule
		\label{tab:dot_unseen}
	\end{tabular}
	\vspace{-6mm}
\end{table}
%

\subsection{Analysis of Learned Features}
\vspace{-1mm}
To demonstrate the effectiveness of the proposed loss, we calculate the similarity (dot product) between the instance-level features of synonymous sentences when jointly training on the RefCOCO, RefCOCO+ and RefCOCOg* datasets.
We randomly sample a pair of synonymous sentences for each image, and compute the average value of all samples in the validation and testing splits of each dataset.
Table~\ref{tab:dot_joint} shows the similarity scores on the RefCOCO, RefCOCO+ and RefCOCOg* datasets, while Table~\ref{tab:dot_unseen} provides the similarity computed on the Ref-Reasoning and ReferItGame datasets in the transfer learning setting.
%
From the results, we observe that the similarity in the embedding space of synonymous sentences is higher when the proposed loss terms are applied in all the settings, which shows that our model is able to transfer the learned features to unseen datasets.

%% file: 5_conclusion.tex
\vspace{-1mm}
\section{Conclusions}
\vspace{-1mm}

In this paper, we focus on the task of referring expression comprehension and tackle the challenge caused by the varieties of synonymous sentences.
To deal with this problem, we propose an end-to-end trainable framework that considers the property in languages for paraphrasing the objects to learn contrastive features.
%
To this end, we employ the feature learning techniques on the image level as well as the instance level to encourage language features describing the same object to attend closely in the visual embedding space, while the expressions identifying different objects to be separated.
We design two negative mining strategies to further facilitate the learning process.
%
Extensive experiments and the ablation study on multiple referring expression datasets demonstrate the effectiveness of the proposed algorithm.
Moreover, in the cross-dataset and transfer learning settings, we show that the proposed method is able to transfer learned representations to other datasets.

%% file: 6_appendix.tex
In this appendix, we provide additional analysis and experimental results, including 1) more implementation details and runtime performance, 2) evaluation on the unseen ReferItGame dataset, and 3) more qualitative results for referring expression comprehension.

\section{Implementation Details and Runtime Performance}
%
The BERT~\cite{Devlin_arxiv_2018} model in our framework is pre-trained on BookCorpus~\cite{Zhu_ICCV_2015} and English Wikipedia.
We use the \textit{BASE} model that has 12 layers of Transformer blocks with each layer having hidden states of size 768 and 12 self-attention heads.
For the GCN in our method, the object proposals are generated from Mask R-CNN~\cite{He_ICCV_2017}. We keep the top $K$ detection candidates for each image, where $K$ is set to 20.
%
%
We then construct a graph $G = (\mathcal{V}, \mathcal{E})$ from the set of object proposals $P = \{p_i\}_{i=1}^K$, where each vertex $v_i \in \mathcal{V}$ corresponds to an object proposal $p_i$, and each edge $e_{ij} \in \mathcal{E}$ models the pairwise relationship between instances $p_i$ and $p_j$.
In the instance-level attention module $A_{ins}$, we compute the word attention on each object proposal to focus on instances that are referred by the sentence.
The word attention $a_i$ on the proposal $p_i$ is defined as the average of all the probabilities that each word $w_t$ refers to $p_i$:
\begin{equation}
a_i = \frac{1}{T} \sum_{t=1}^T s_{i,t} = \frac{1}{T} \sum_{t=1}^T \langle f_{w_t}, f_{p_i} \rangle,
\label{eq:word}
\end{equation}
where $T$ is the number of words in the sentence, and $s_{i,t}$ is the inner product between the feature $f_{w_t}$ of word $w_t$ and the average pooled feature $f_{p_i}$ of proposal $p_i$.
To compute the feature node $f_{v_i}$ at vertex $v_i$, we first concatenate the average pooled feature $f_{p_i}$ and the 5-dimensional location feature~\cite{Mao_CVPR_2016} $[\frac{x_{tl}}{W}, \frac{y_{tl}}{H}, \frac{x_{br}}{W}, \frac{y_{br}}{H}, \frac{wh}{WH}]$ on proposal $p_i$, where $(x_{tl}, y_{tl})$ and $(x_{br}, y_{br})$ are the coordinates of the top-left and bottom-right corners of the proposal, $h$ and $w$ are height and width of the proposal, $H$ and $W$ are height and width of the image.
Then, the concatenated feature is multiplied by the word attention $a_i$ in (\ref{eq:word}) to form $f_{v_i}$.
%
We use a two-layer GCN to capture second-order interactions.

Our framework is implemented on a machine with an Intel Xeon 2.3 GHz processor and an NVIDIA GTX 1080 Ti GPU with 11 GB memory.
We report the runtime performance in Table~\ref{tab:runtime}. Our end-to-end model takes 0.325 seconds to process one image in a single forward pass, which is much faster than MAttNet~\cite{Yu_CVPR_2018} and CM-Att-Erase~\cite{Liu_CVPR_2019} that require multiple steps of inference. All the source codes and models will be made available to the public.
\begin{table}[t]
	\centering
	\small
	\caption{\textbf{Runtime performance.}}
	\begin{tabular}{cc}
		\toprule
		Method & Time (s) \\
		\midrule
		MAttNet~\cite{Yu_CVPR_2018} & 0.671 \\
		CM-Att-Erase~\cite{Liu_CVPR_2019} & 0.734 \\
        Ours & 0.325 \\
		\bottomrule
		\label{tab:runtime}
	\end{tabular}
	\vspace{-6mm}
\end{table}

\section{Ablation Study}
In Table~\ref{tab:ablation_bert}, we provide more ablation study on the language encoder in our framework. The models are jointly trained on the RefCOCO, RefCOCO+ and RefCOCOg* datasets.
We show the results of the baseline model without our loss and the full model using the Bi-LSTM or BERT as the language encoder.
The results demonstrate that the performance gains against the baseline are mainly contributed by the proposed image- and instance-level loss (row 1 vs. row 3) rather than the BERT model (row 2 vs. row 3).

\begin{table*}[t]
	\centering
    \small
	\caption{\textbf{Ablation study of jointly training on three datasets.} We show that the performance gains mainly come from the proposed loss terms rather than the BERT model.}
	\vspace{1mm}
	\begin{tabular}{cccccccccccc}
		\toprule
		& & & \multicolumn{3}{c}{RefCOCO} & \multicolumn{3}{c}{RefCOCO+} & \multicolumn{3}{c}{RefCOCOg*} \\ 
		\cmidrule(lr){4-6} \cmidrule(lr){7-9} \cmidrule(lr){10-12}
		Method & Our Loss & Language Encoder & val & testA & testB & val & testA & testB & val & testA & testB \\ 
		\midrule
		Baseline &  & BERT & 75.17 & 79.53 & 68.72 & 63.42 & 70.55 & 53.38 & 60.62 & 64.57 & 53.81 \\
		Final w/ Bi-LSTM & \checkmark & Bi-LSTM & 81.37 & 84.58 & 74.39 & 69.55 & 76.83 & 60.66 & 68.73 & 70.89 & 61.35 \\
		Final w/ BERT & \checkmark & BERT & \textbf{82.42} & \textbf{85.77} & \textbf{75.29} & \textbf{70.64} & \textbf{78.12} & \textbf{61.49} & \textbf{69.92} & \textbf{71.75} & \textbf{62.68} \\
		
		\bottomrule
		\label{tab:ablation_bert}
	\end{tabular}
	\vspace{-4mm}
\end{table*}

\section{More Evaluation on the Unseen Dataset}
To demonstrate the ability of our model to transfer learned features to unseen datasets, we use the models trained under various settings to evaluate on the ReferItGame \cite{Kazemzadeh_EMNLP_2014} dataset, which consists of completely different images and expressions from our training datasets.
The results are presented in Table~\ref{tab:unseen_referitgame}, and the performance of our model trained on the ReferItGame dataset under the fully supervised setting is provided for reference.
We first show the results of models trained on the RefCOCOg dataset. When training with the proposed intra-dataset loss, the performance is improved from the baseline and better than the two-stage MAttNet~\cite{Yu_CVPR_2018} and CM-Att-Erase~\cite{Liu_CVPR_2019} approaches. Note that all the models use the same ResNet-101 backbone as the feature extractor.
Then we jointly train our models on the RefCOCO, RefCOCO+ and RefCOCOg* (our split) datasets. When more datasets are included to train our full model, we observe more performance gain over the baseline, which demonstrates the effectiveness of our feature learning technique considering synonymous sentences.

\begin{table}[t]
	\centering
	\small
	\vspace{-125mm}
	\caption{\textbf{Evaluation on the ReferItGame dataset with models trained on different datasets.} The models are trained on the ReferItGame (top group), RefCOCOg (middle group), or RefCOCO, RefCOCO+ and RefCOCOg* datasets (bottom group).}
	\begin{tabular}{ccc}
		\toprule
		& & Split \\ \cmidrule(l){3-3} 
		Method & Training Dataset & test \\ \midrule
		
		Ours (supervised) & ReferItGame & 57.04 \\
		
		\midrule
		MAttNet~\cite{Yu_CVPR_2018}       & \multirow{4}{*}{RefCOCOg}     & 42.67 \\
		CM-Att-Erase~\cite{Liu_CVPR_2019} &                               & 43.39 \\
		Ours (baseline)   &                                               & 43.18 \\
		Ours (full model) &                                               & \textbf{45.72} \\ \midrule
		Ours (baseline)   & \multirow{2}{*}{\makecell{RefCOCO, RefCOCO+,\\RefCOCOg*}} & 45.53 \\
		Ours (full model) &                                               & \textbf{48.21} \\
		\bottomrule
		\label{tab:unseen_referitgame}
	\end{tabular}
	\vspace{-4mm}
\end{table}

\section{Qualitative Results}
\begin{figure*}[t]
	\centering
	\footnotesize
	\begin{tabular}
		{ @{\hspace{0mm}}c@{\hspace{1mm}} @{\hspace{0mm}}c@{\hspace{1mm}} @{\hspace{0mm}}c@{\hspace{1mm}} @{\hspace{0mm}}c@{\hspace{0mm}}}
		\makecell{``left elephant''} & \makecell{``elephant behind the others''} & \makecell{``far right elephant''} & \makecell{``elephant most in the water''} \\
		\includegraphics[width=0.24\linewidth]{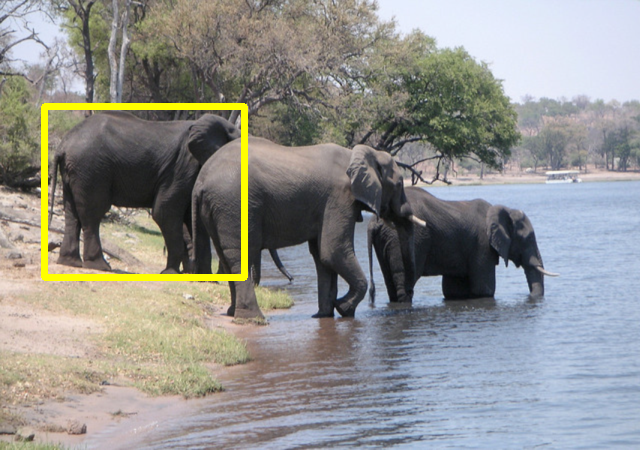} &
		\includegraphics[width=0.24\linewidth]{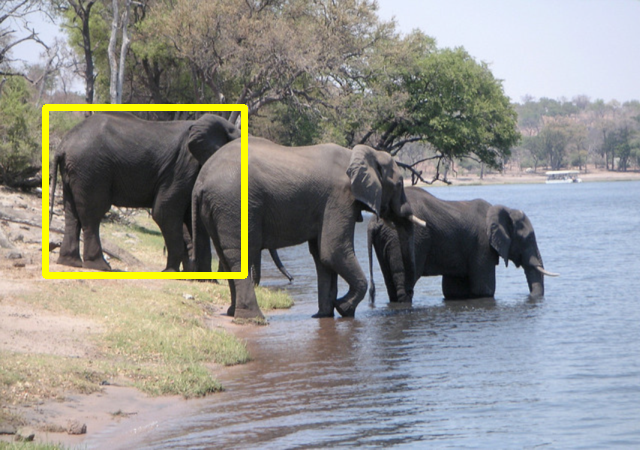} &
		\includegraphics[width=0.24\linewidth]{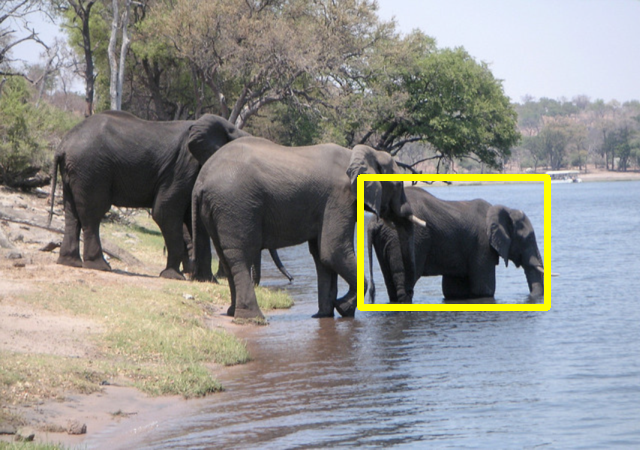} &
		\includegraphics[width=0.24\linewidth]{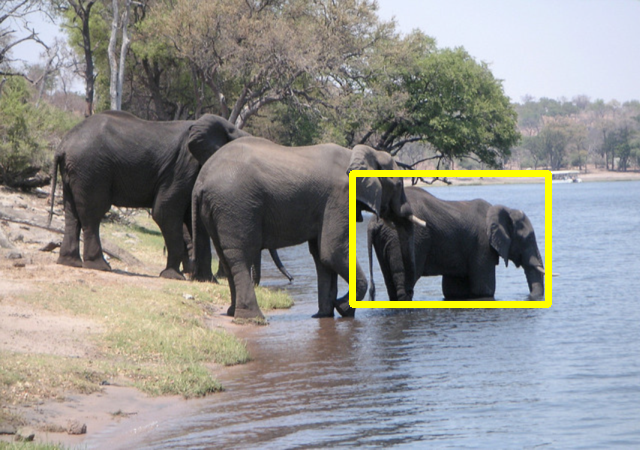} \\
		\makecell{``kid in black''} & \makecell{``kid skiing''} & \makecell{``woman in blue''} & \makecell{``skier in light blue jacket''} \\
		\includegraphics[width=0.24\linewidth]{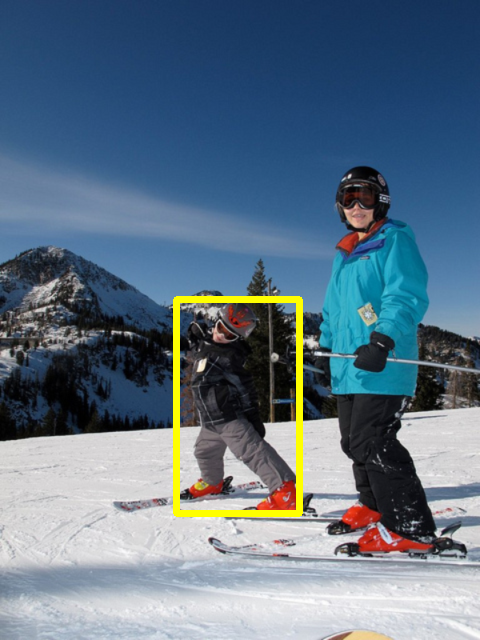} &
		\includegraphics[width=0.24\linewidth]{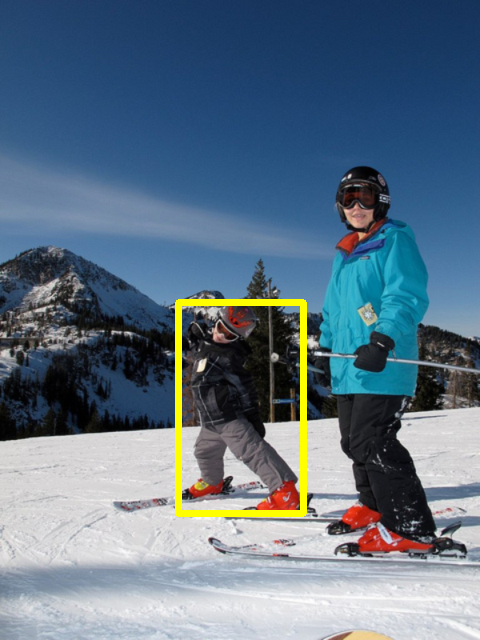} &
		\includegraphics[width=0.24\linewidth]{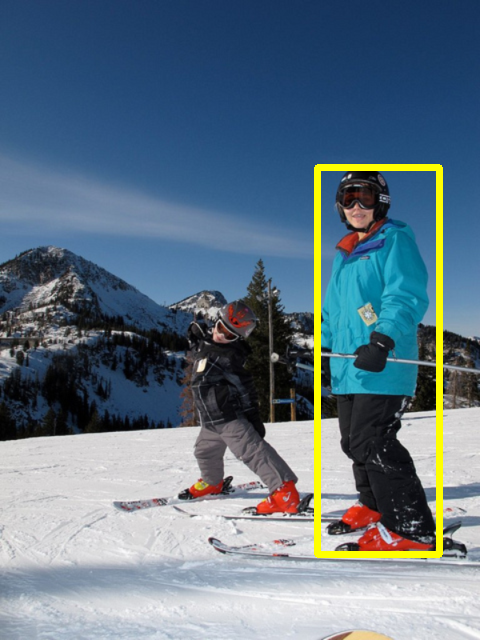} &
		\includegraphics[width=0.24\linewidth]{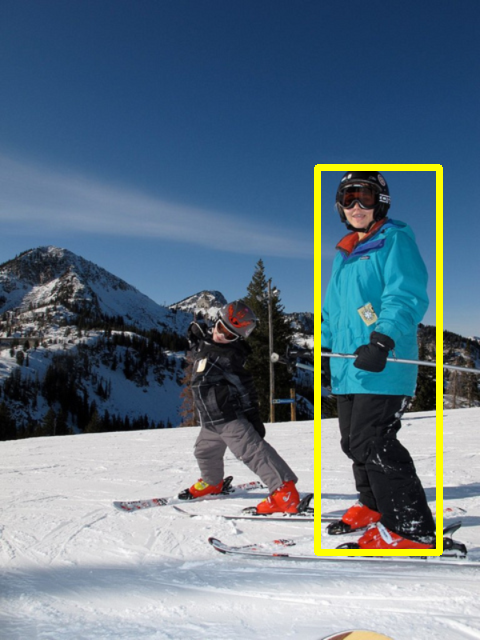} \\
		\makecell{``woman on left''} & \makecell{``lady in black''} & \makecell{``man in front to the right''} & \makecell{``green striped shirt''} \\
		\includegraphics[width=0.24\linewidth]{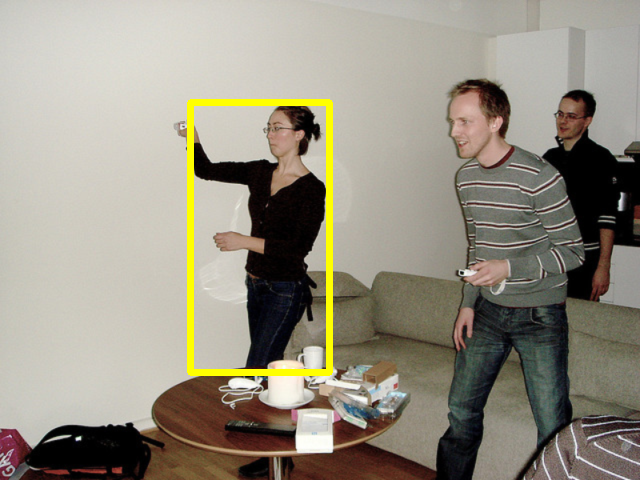} &
		\includegraphics[width=0.24\linewidth]{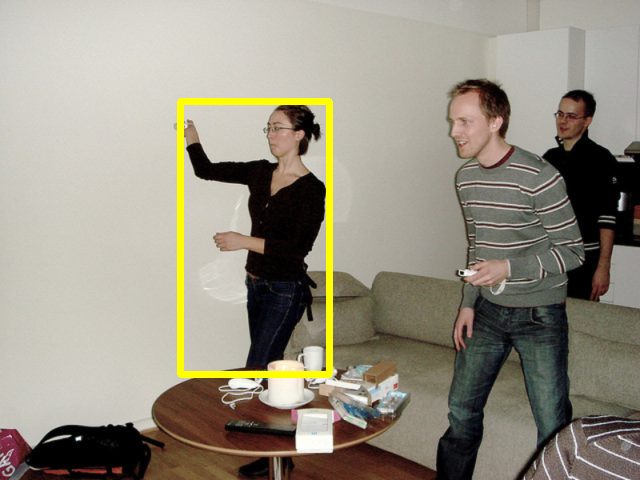} &
		\includegraphics[width=0.24\linewidth]{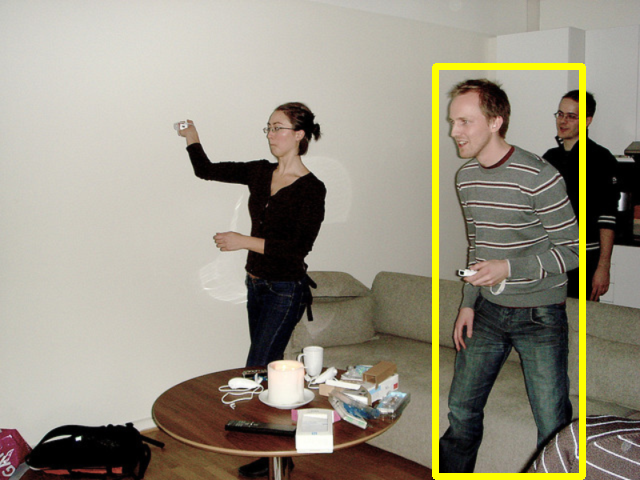} &
		\includegraphics[width=0.24\linewidth]{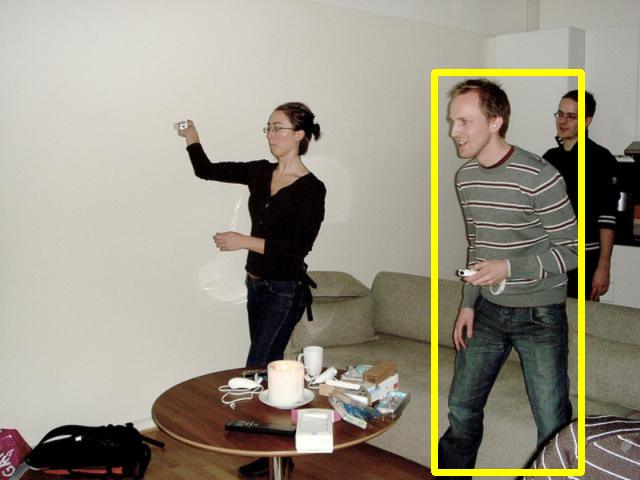} \\
		\makecell{``girl with dog''} & \makecell{``black shirt''} & \makecell{``woman bending''} & \makecell{``pink skirt''} \\
		\includegraphics[width=0.24\linewidth]{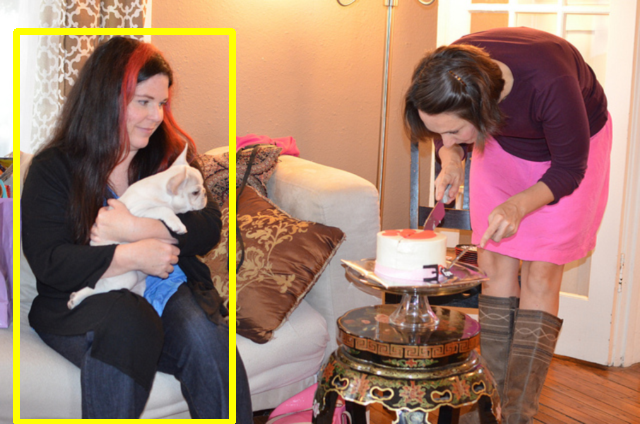} &
		\includegraphics[width=0.24\linewidth]{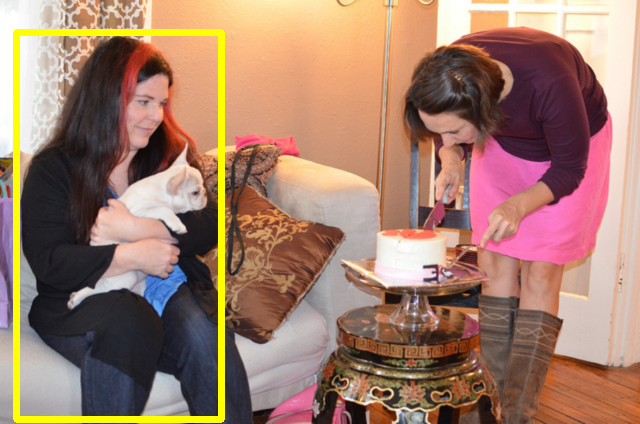} &
		\includegraphics[width=0.24\linewidth]{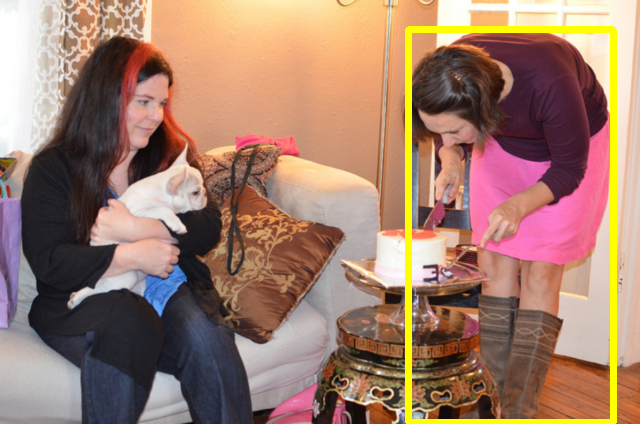} &
		\includegraphics[width=0.24\linewidth]{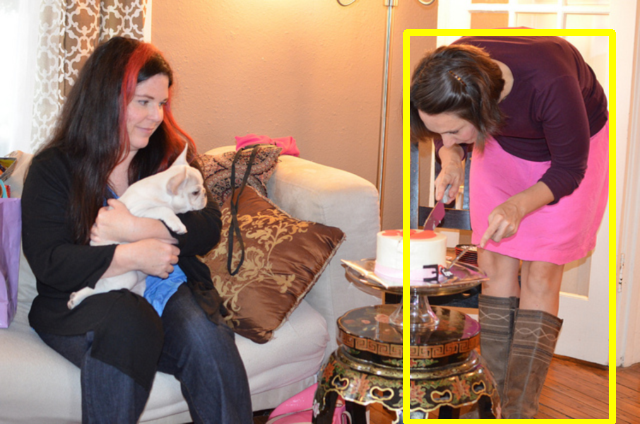} \\
		\makecell{``right van''} & \makecell{``mini van''} & \makecell{``white car in front of bus''} & \makecell{``old white car''} \\
		\includegraphics[width=0.24\linewidth]{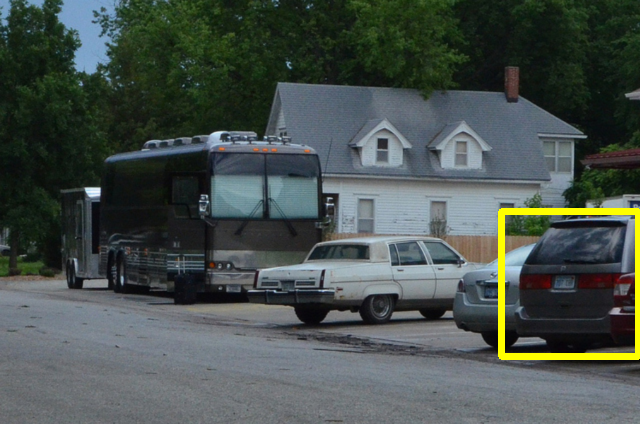} &
		\includegraphics[width=0.24\linewidth]{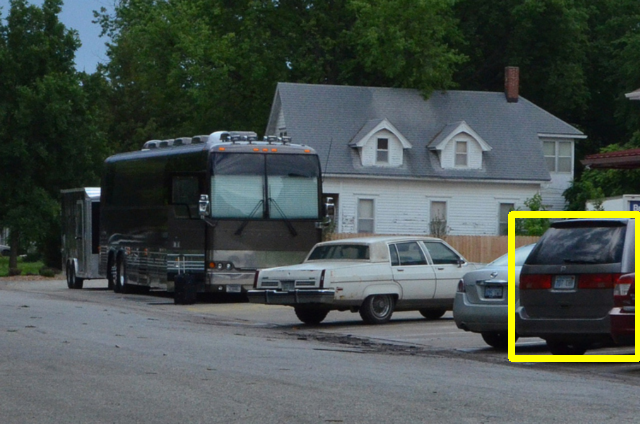} &
		\includegraphics[width=0.24\linewidth]{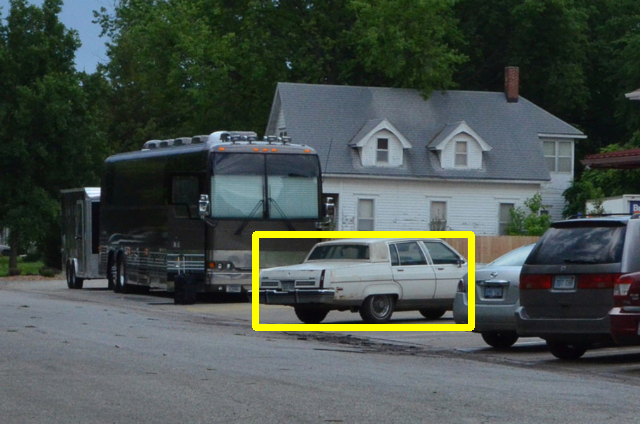} &
		\includegraphics[width=0.24\linewidth]{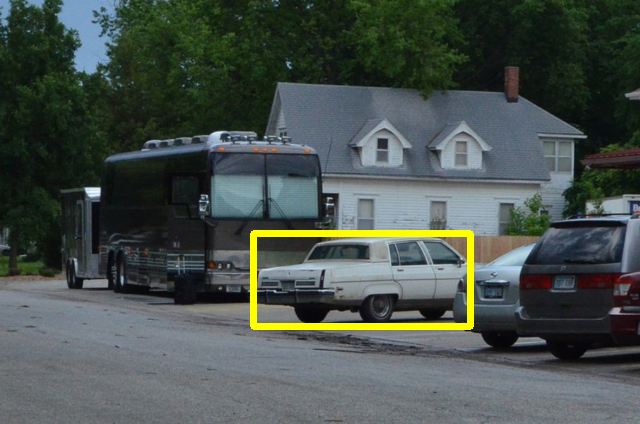} \\
		\makecell{``guy''} & \makecell{``blue shirt''} & \makecell{``animal on right''} & \makecell{``the standing cow''} \\
		\includegraphics[width=0.24\linewidth]{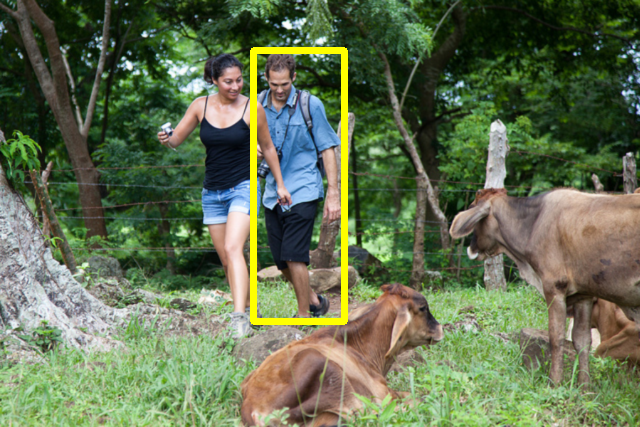} &
		\includegraphics[width=0.24\linewidth]{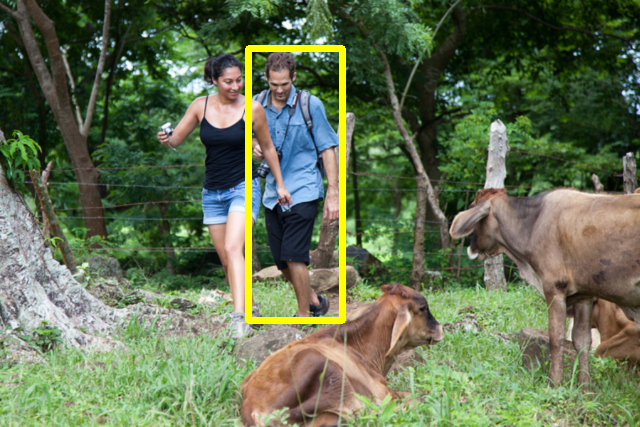} &
		\includegraphics[width=0.24\linewidth]{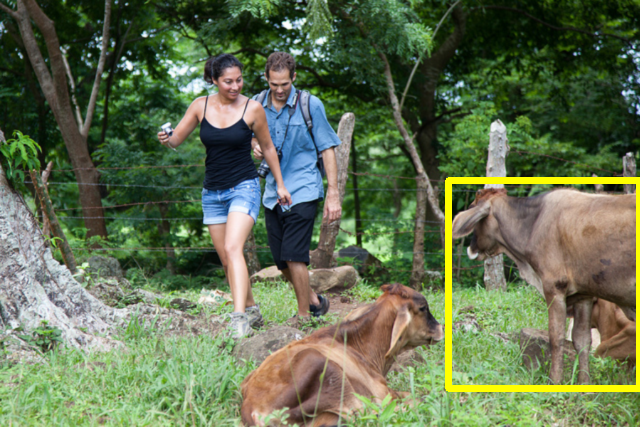} &
		\includegraphics[width=0.24\linewidth]{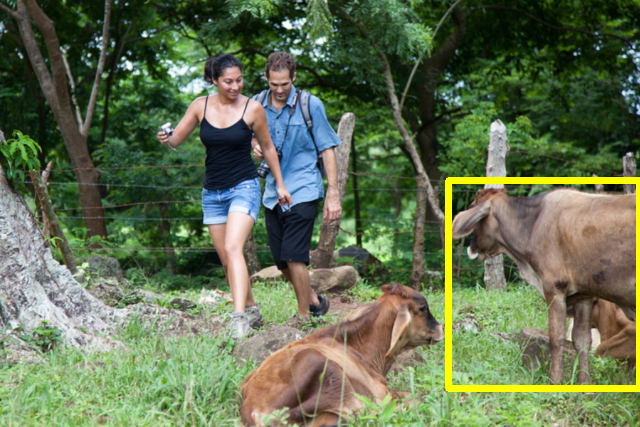} \\
	\end{tabular}
	\caption{\textbf{Sample results of jointly training on the RefCOCO, RefCOCO+ and RefCOCOg* datasets.}}
	\label{fig:final_1}
\end{figure*}

\begin{figure*}[t]
	\centering
	\footnotesize
	\begin{tabular}
		{ @{\hspace{0mm}}c@{\hspace{1mm}} @{\hspace{0mm}}c@{\hspace{1mm}} @{\hspace{0mm}}c@{\hspace{1mm}} @{\hspace{0mm}}c@{\hspace{0mm}}}
		\makecell{``left baby''} & \makecell{``boy in plaid''} & \makecell{``right kid''} & \makecell{``strip shirt boy eyes closed''} \\
		\includegraphics[width=0.24\linewidth]{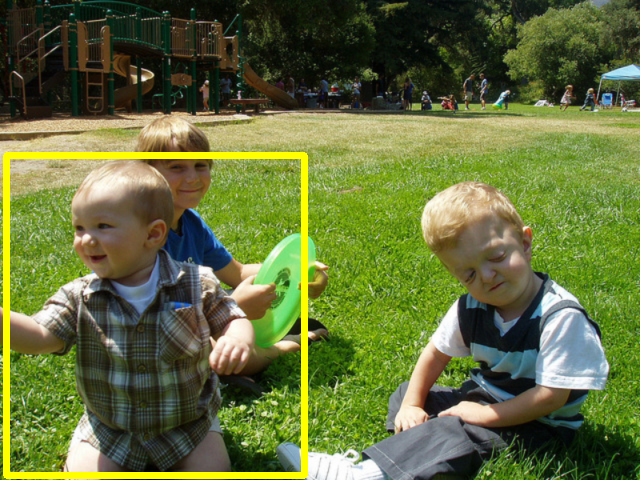} &
		\includegraphics[width=0.24\linewidth]{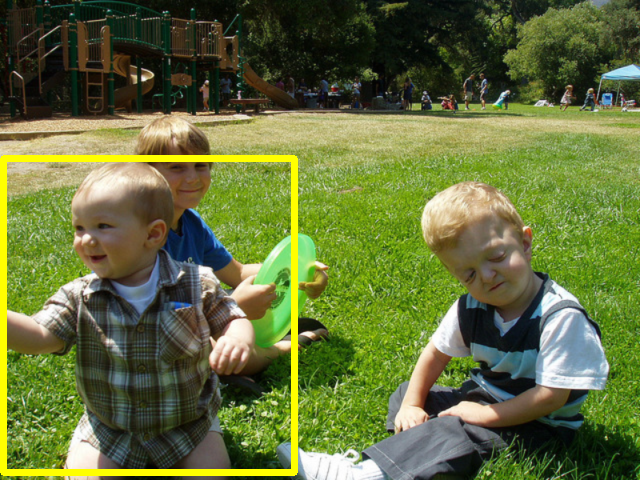} &
		\includegraphics[width=0.24\linewidth]{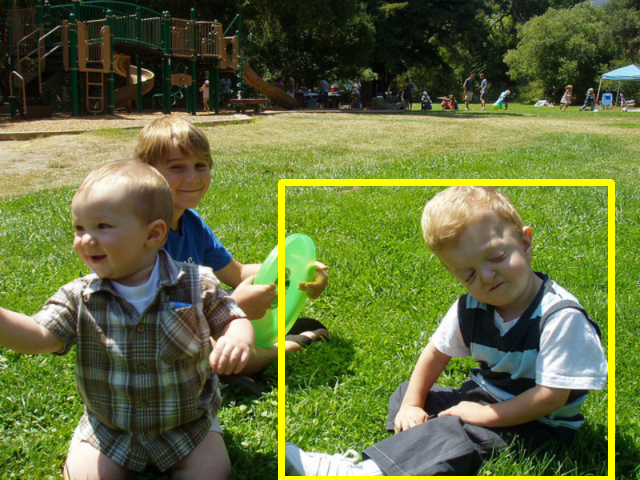} &
		\includegraphics[width=0.24\linewidth]{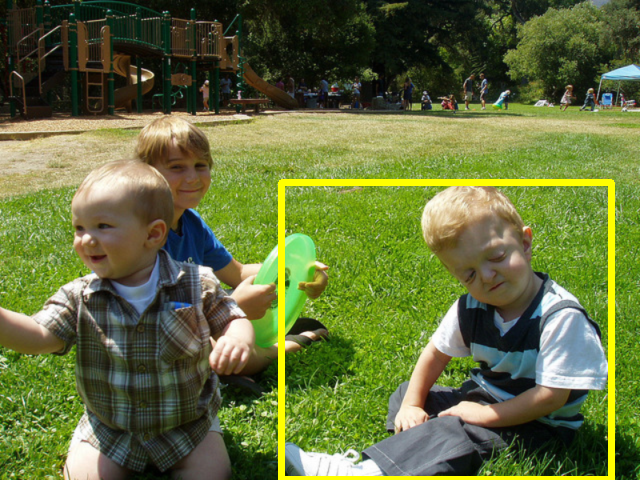} \\
		\makecell{``right couch with cat on it''} & \makecell{``furthest couch''} & \makecell{``bottom sofa''} & \makecell{``sofa closest''} \\
		\includegraphics[width=0.24\linewidth]{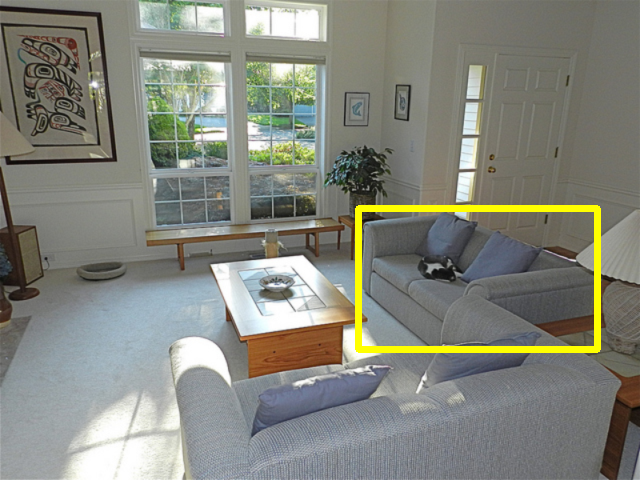} &
		\includegraphics[width=0.24\linewidth]{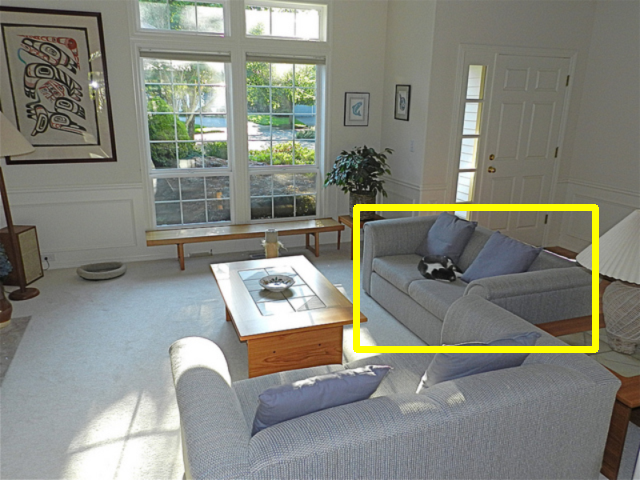} &
		\includegraphics[width=0.24\linewidth]{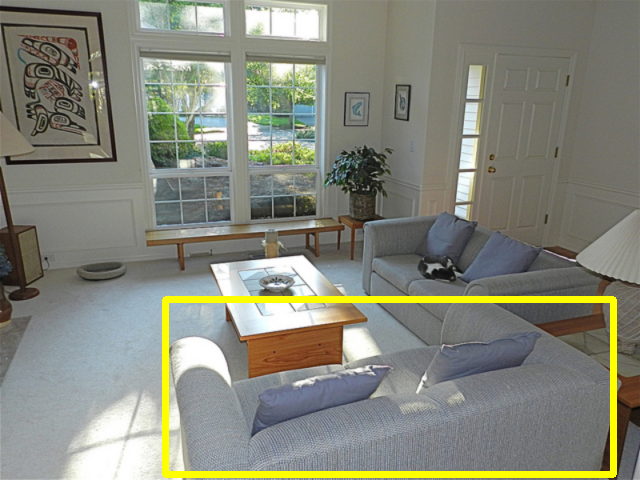} &
		\includegraphics[width=0.24\linewidth]{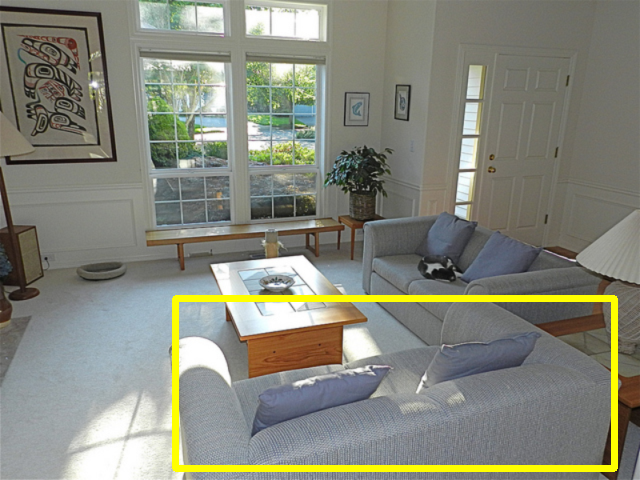} \\
		\makecell{``man farthest right''} & \makecell{``man with gray coat''} & \makecell{``woman middle front white jacket''} & \makecell{``white jacket lady off steps''} \\
		\includegraphics[width=0.24\linewidth]{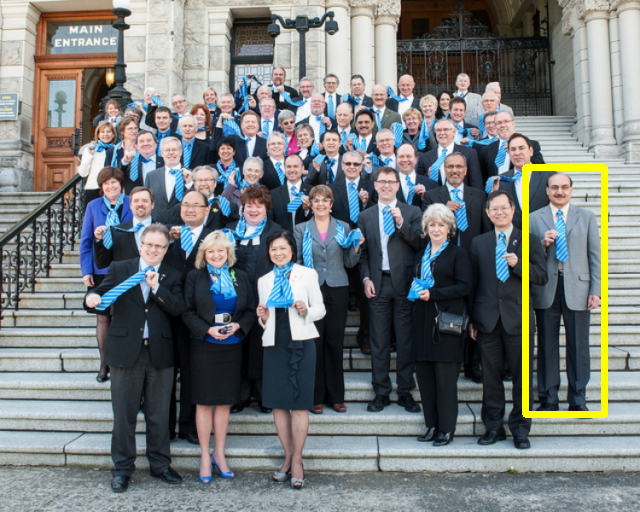} &
		\includegraphics[width=0.24\linewidth]{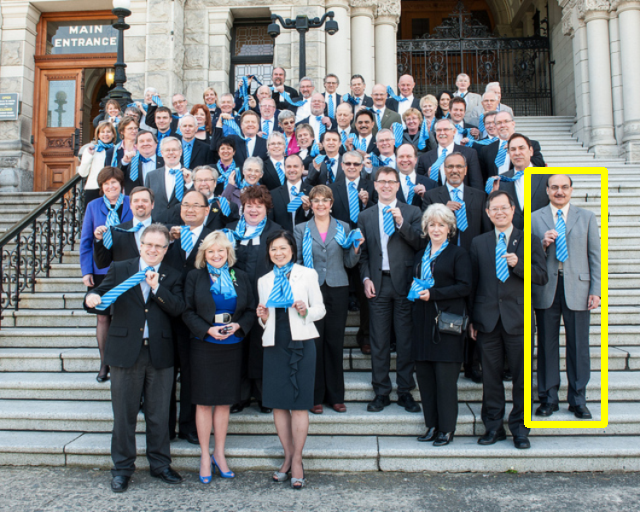} &
		\includegraphics[width=0.24\linewidth]{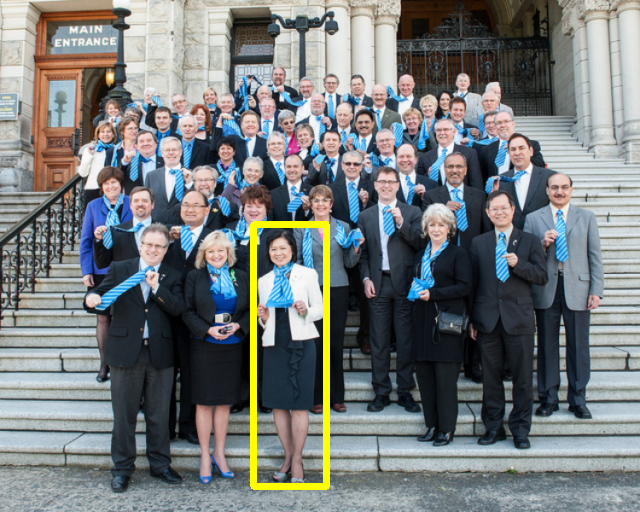} &
		\includegraphics[width=0.24\linewidth]{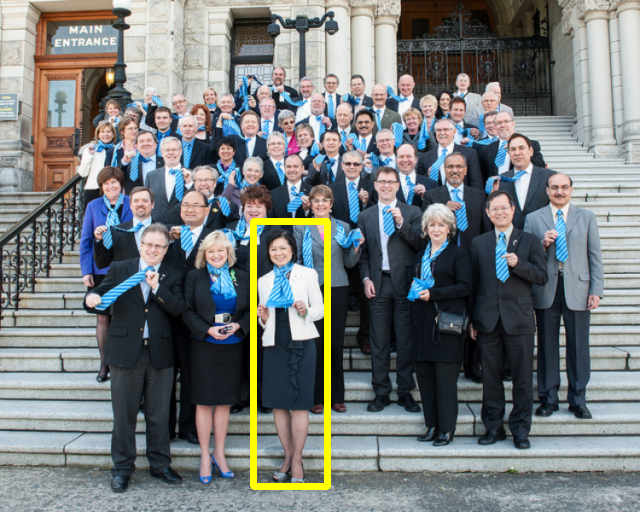} \\
		\makecell{``boy on left''} & \makecell{``kid under purple umbrella''} & \makecell{``child on right''} & \makecell{``child holding orange umbrella''} \\
		\includegraphics[width=0.24\linewidth]{figure/supp/01191_COCO_train2014_000000452619_0_box_pred.png} &
		\includegraphics[width=0.24\linewidth]{figure/supp/01191_COCO_train2014_000000452619_1_box_pred.png} &
		\includegraphics[width=0.24\linewidth]{figure/supp/01191_COCO_train2014_000000452619_3_box_pred.png} &
		\includegraphics[width=0.24\linewidth]{figure/supp/01191_COCO_train2014_000000452619_4_box_pred.png} \\
		\makecell{``center lady white shirt''} & \makecell{``female watching''} & \makecell{``man in white shirt\\in front of white poster board''} & \makecell{``guy swinging''} \\
		\includegraphics[width=0.24\linewidth]{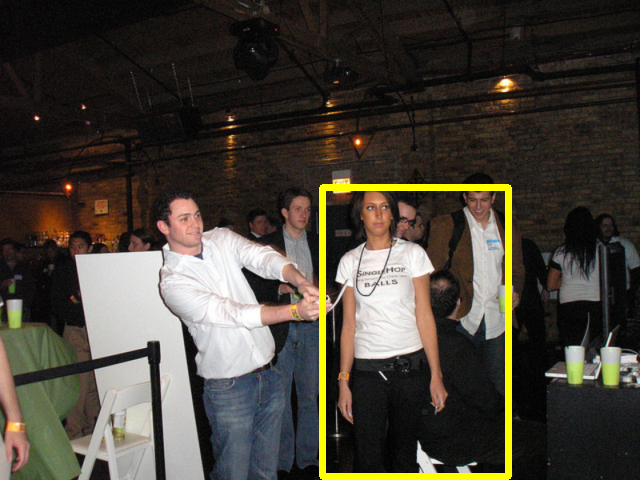} &
		\includegraphics[width=0.24\linewidth]{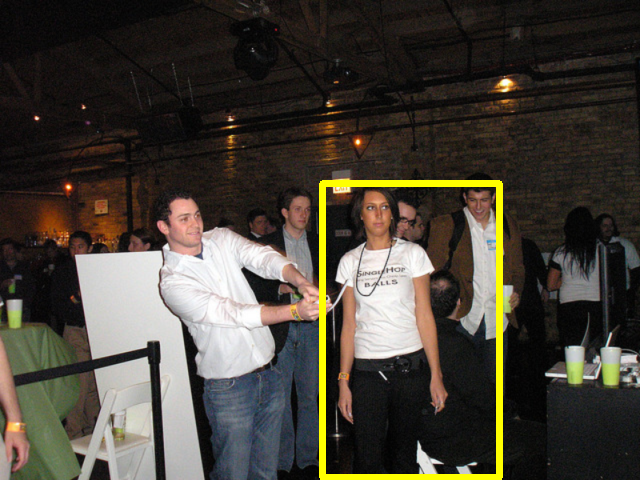} &
		\includegraphics[width=0.24\linewidth]{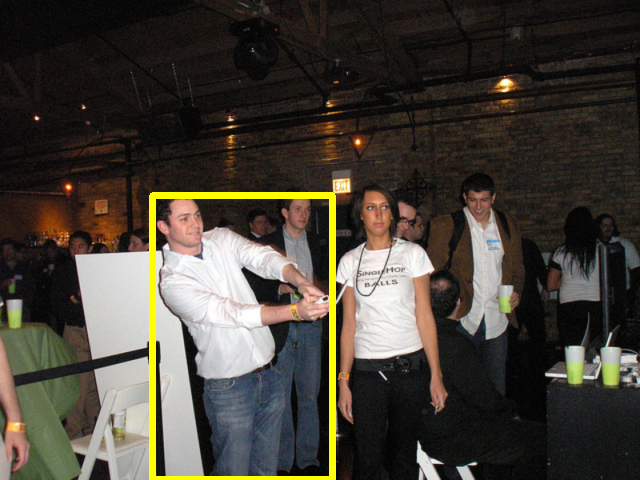} &
		\includegraphics[width=0.24\linewidth]{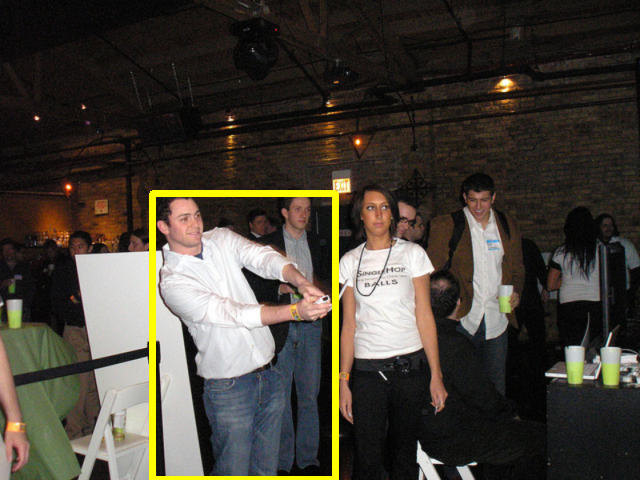} \\
		\makecell{``donut in middle on left side''} & \makecell{``donut above sprinkles''} & \makecell{``chocolate donut\\second from right in front row''} & \makecell{``nearest chocolate donut''} \\
		\includegraphics[width=0.24\linewidth]{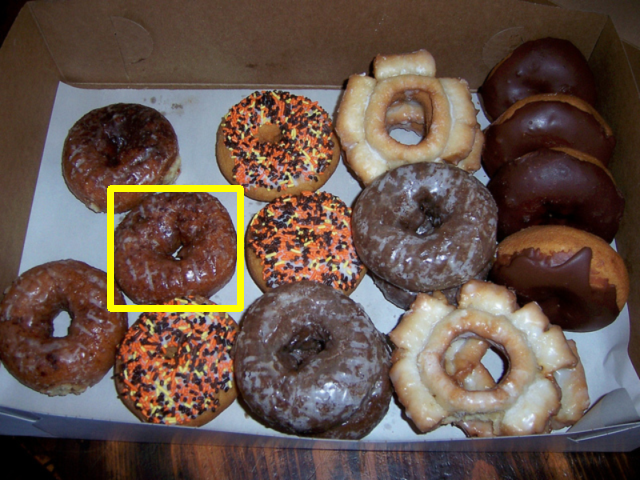} &
		\includegraphics[width=0.24\linewidth]{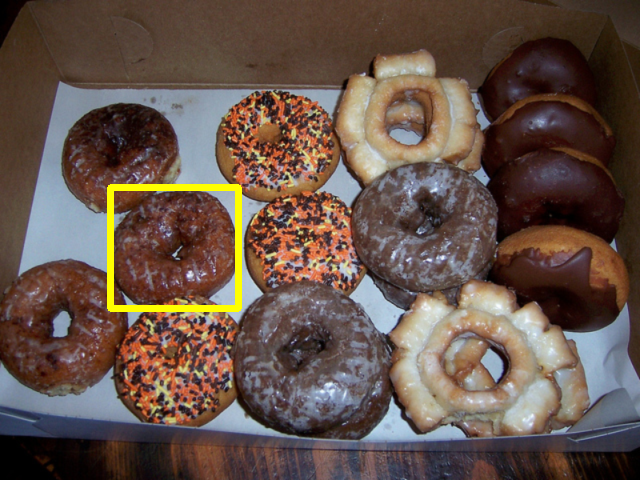} &
		\includegraphics[width=0.24\linewidth]{figure/supp/01454_COCO_train2014_000000425063_8_box_pred.png} &
		\includegraphics[width=0.24\linewidth]{figure/supp/01454_COCO_train2014_000000425063_9_box_pred.png} \\
	\end{tabular}
	\caption{\textbf{Sample results of jointly training on the RefCOCO, RefCOCO+ and RefCOCOg* datasets.}}
	\label{fig:final_2}
\end{figure*}

\begin{figure*}[t]
	\centering
	\footnotesize
	\begin{tabular}
		{ @{\hspace{0mm}}c@{\hspace{1mm}} @{\hspace{0mm}}c@{\hspace{1mm}} @{\hspace{0mm}}c@{\hspace{1mm}} @{\hspace{0mm}}c@{\hspace{0mm}}}
		\makecell{``girl right standing''} & \makecell{``woman handing cake''} & \makecell{``the young child on the left\\sitting near the old''} & \makecell{``little girl looking up''} \\
		\includegraphics[width=0.24\linewidth]{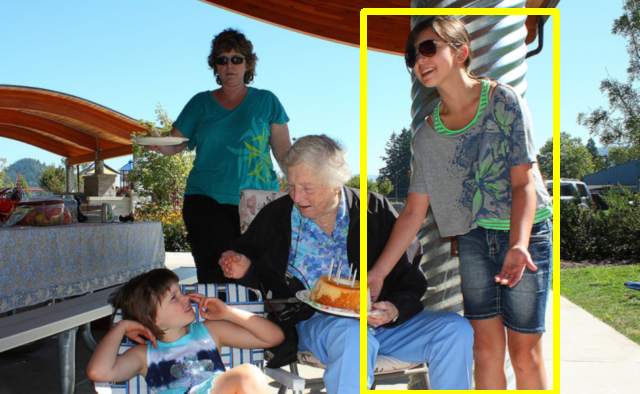} &
		\includegraphics[width=0.24\linewidth]{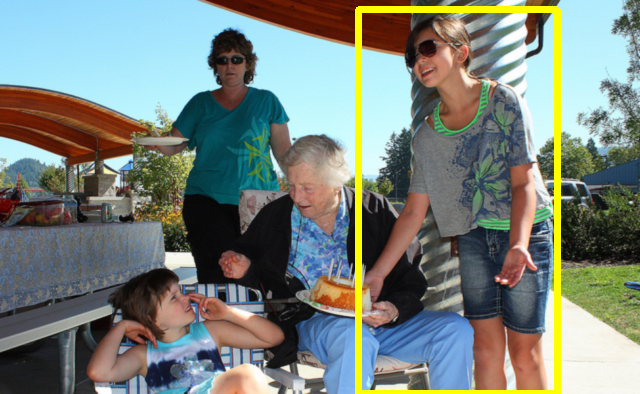} &
		\includegraphics[width=0.24\linewidth]{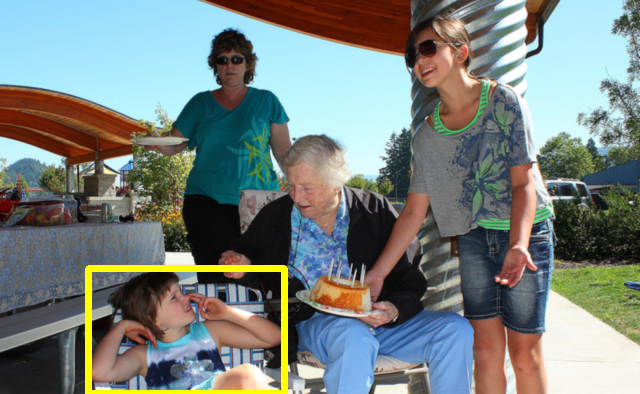} &
		\includegraphics[width=0.24\linewidth]{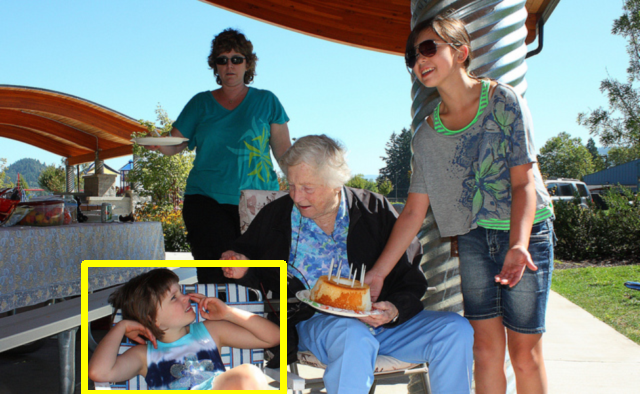} \\
		\makecell{``second horse from the left''} & \makecell{``black horse next to white''} & \makecell{``white horse on left''} & \makecell{``white horse with head down''} \\
		\includegraphics[width=0.24\linewidth]{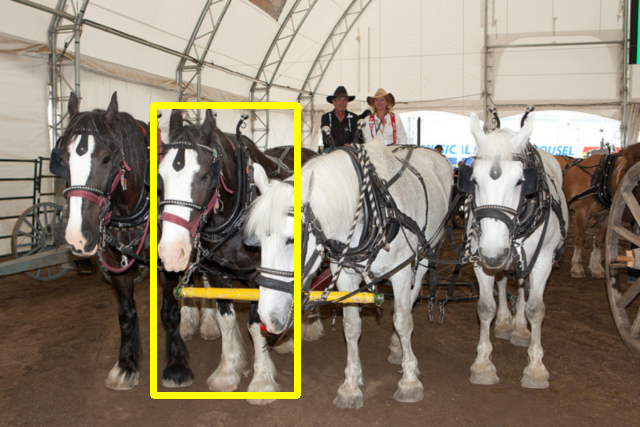} &
		\includegraphics[width=0.24\linewidth]{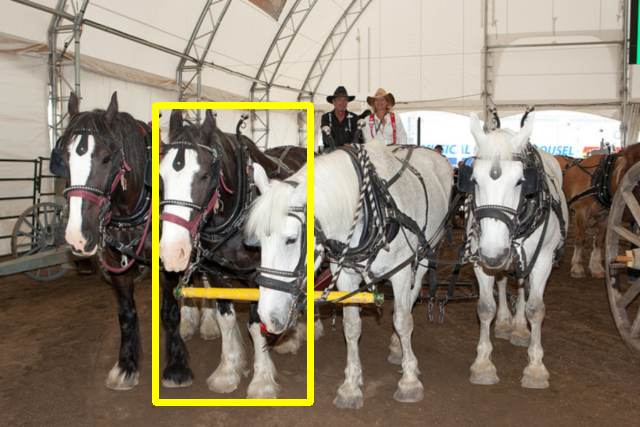} &
		\includegraphics[width=0.24\linewidth]{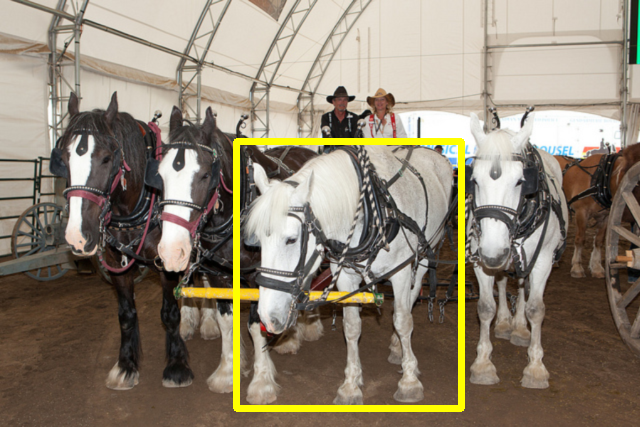} &
		\includegraphics[width=0.24\linewidth]{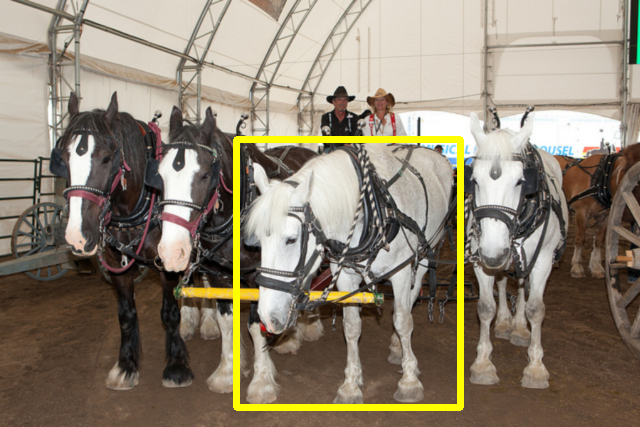} \\
		\makecell{``the blue jacket dude on the right''} & \makecell{``man in blue shirt''} & \makecell{``guy with gray sweater''} & \makecell{``man with back to camera''} \\
		\includegraphics[width=0.24\linewidth]{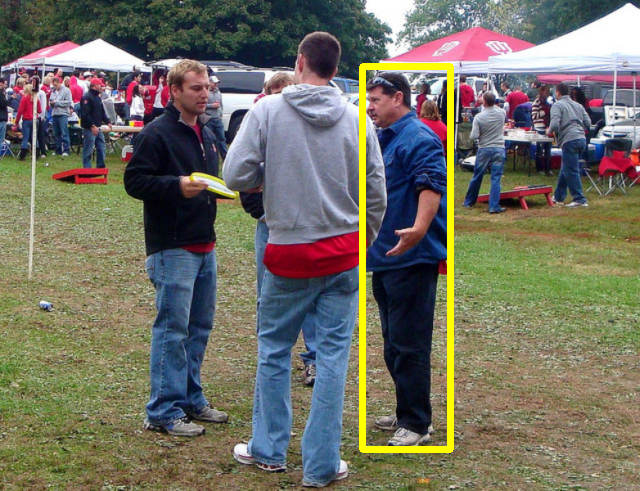} &
		\includegraphics[width=0.24\linewidth]{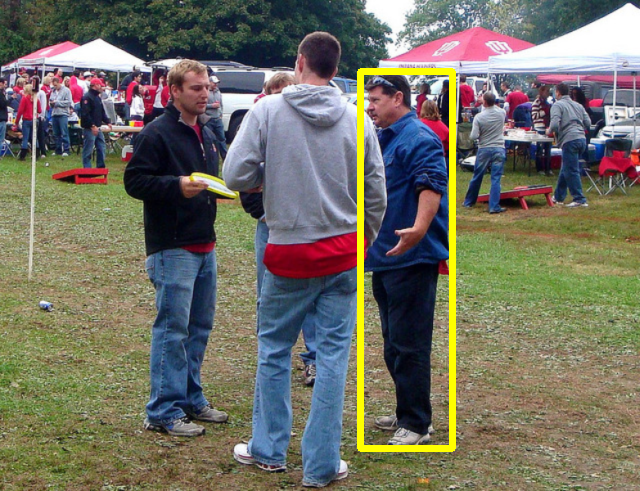} &
		\includegraphics[width=0.24\linewidth]{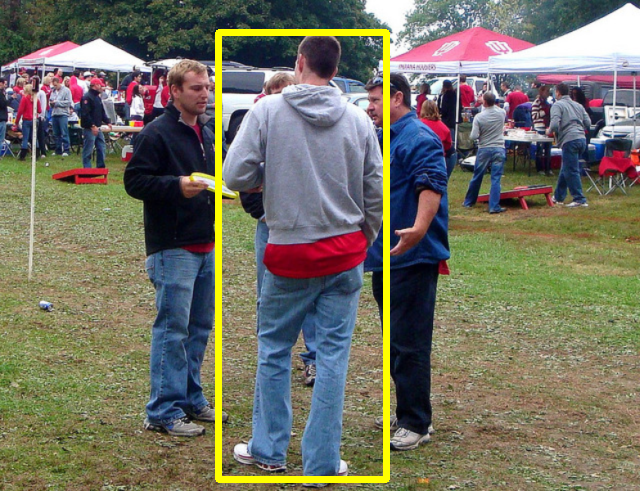} &
		\includegraphics[width=0.24\linewidth]{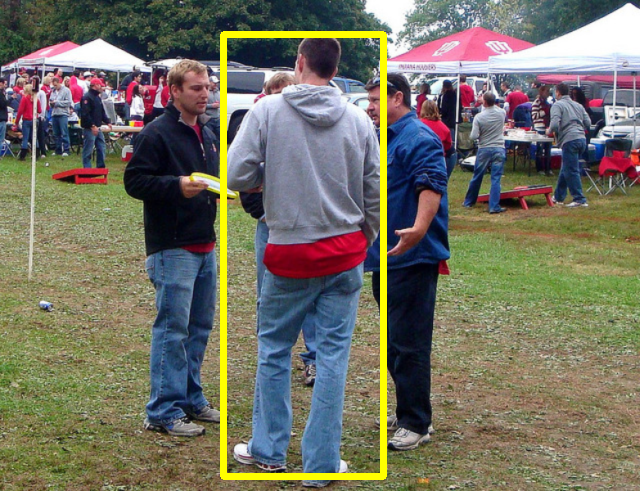} \\
		\makecell{``zebra on far right''} & \makecell{``zebra alone''} & \makecell{``left zebra''} & \makecell{``zebra closest''} \\
		\includegraphics[width=0.24\linewidth]{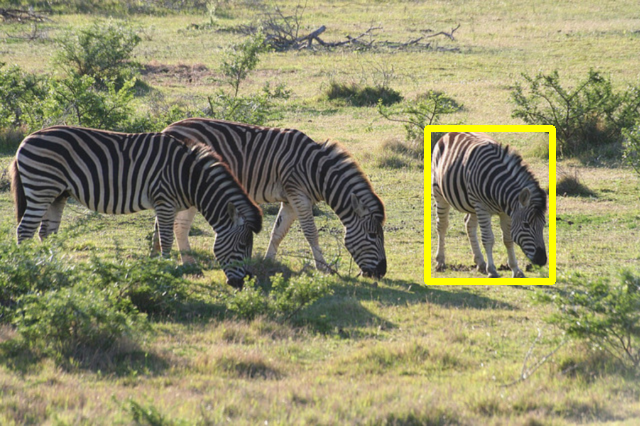} &
		\includegraphics[width=0.24\linewidth]{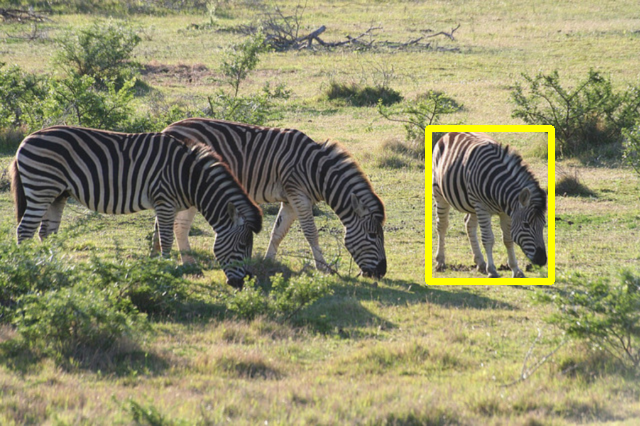} &
		\includegraphics[width=0.24\linewidth]{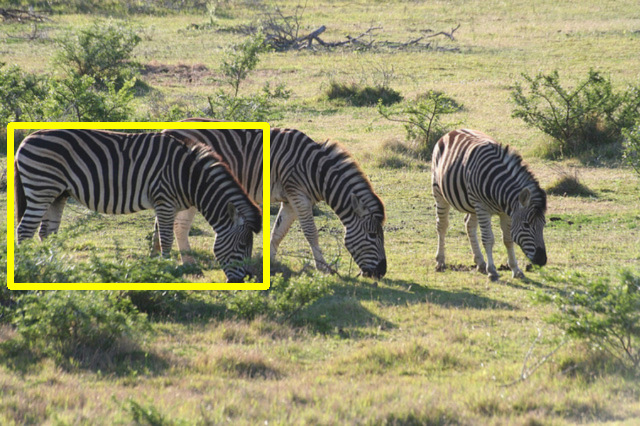} &
		\includegraphics[width=0.24\linewidth]{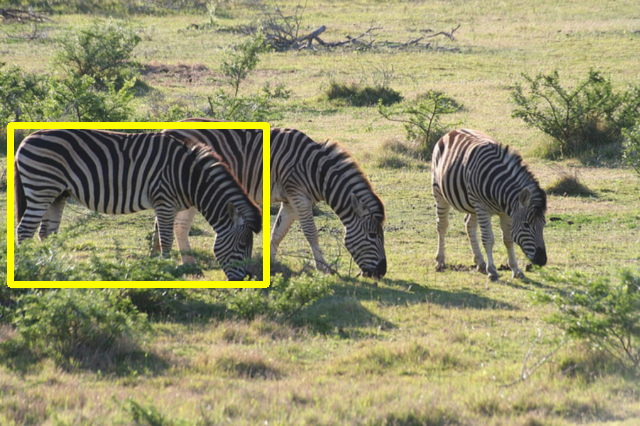} \\
		\makecell{``rightmost person''} & \makecell{``baseball player''} & \makecell{``man in red''} & \makecell{``man wearing red shirt\\beside woman''} \\
		\includegraphics[width=0.24\linewidth]{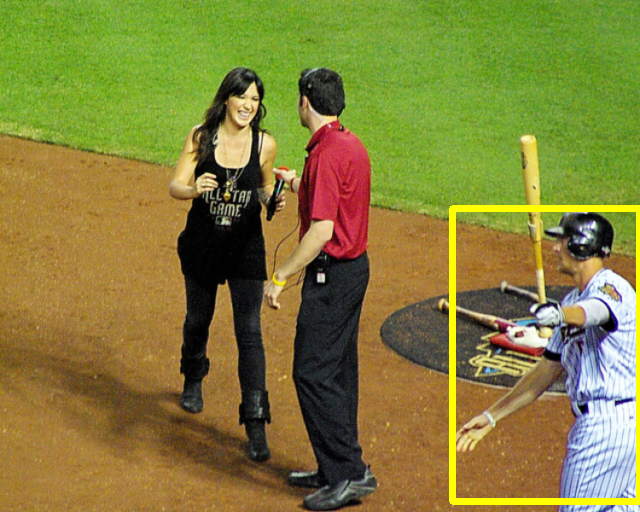} &
		\includegraphics[width=0.24\linewidth]{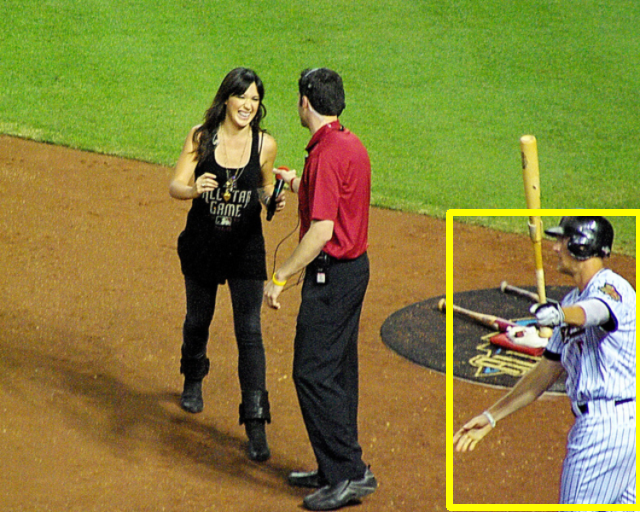} &
		\includegraphics[width=0.24\linewidth]{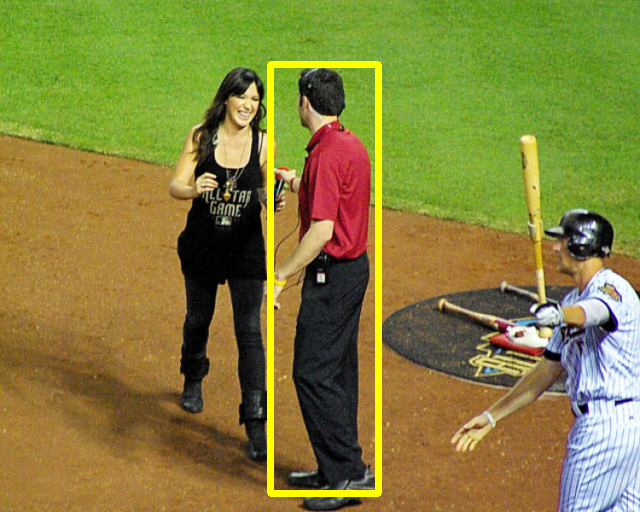} &
		\includegraphics[width=0.24\linewidth]{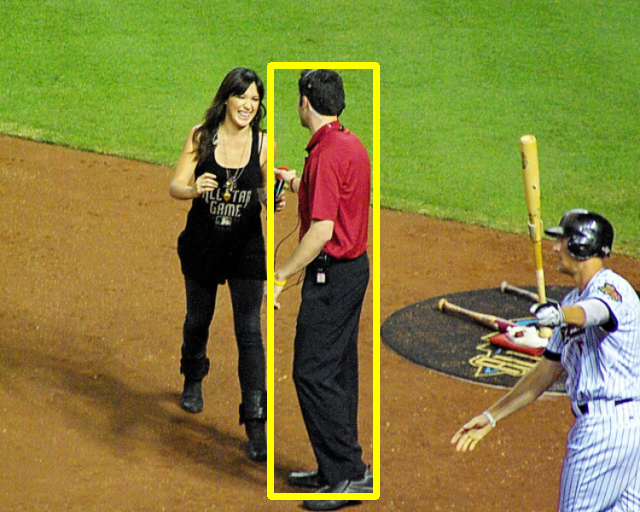} \\
	\end{tabular}
	\caption{\textbf{Sample results of jointly training on the RefCOCO, RefCOCO+ and RefCOCOg* datasets.}}
	\label{fig:final_3}
\end{figure*}

\begin{figure*}[t]
	\centering
	\footnotesize
	\begin{tabular}
		{ @{\hspace{0mm}}c@{\hspace{1mm}} @{\hspace{0mm}}c@{\hspace{1mm}} @{\hspace{0mm}}c@{\hspace{1mm}} @{\hspace{0mm}}c@{\hspace{0mm}}}
		\makecell{``man in white with black hair''} & \makecell{``middle orange''} & \makecell{``the sheep behind\\the sheep looking at the camera''} & \makecell{``hands''} \\
		\includegraphics[height=0.17\linewidth]{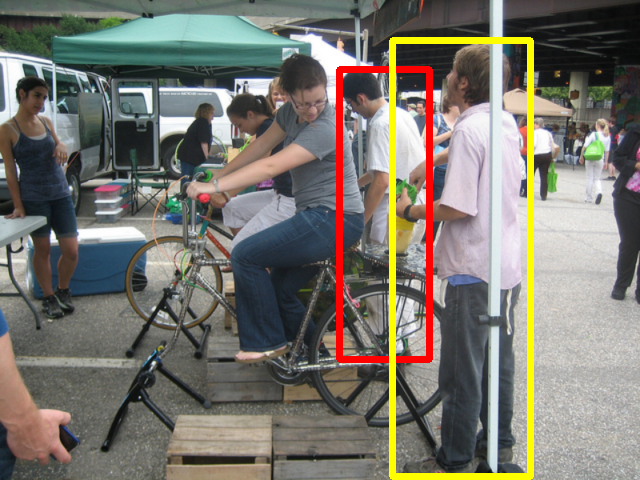} &
		\includegraphics[height=0.17\linewidth]{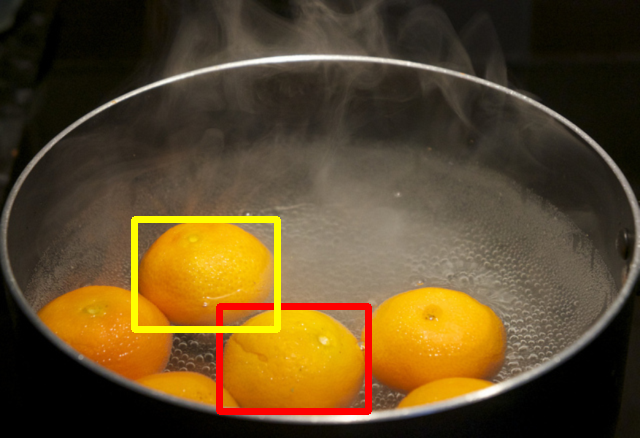} &
		\includegraphics[height=0.17\linewidth]{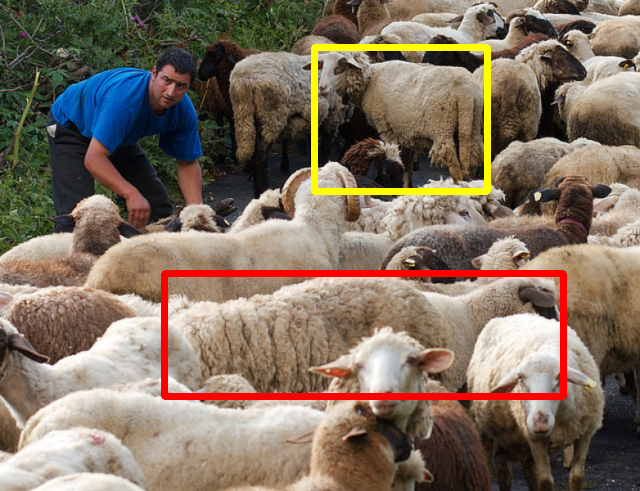} &
		\includegraphics[height=0.17\linewidth]{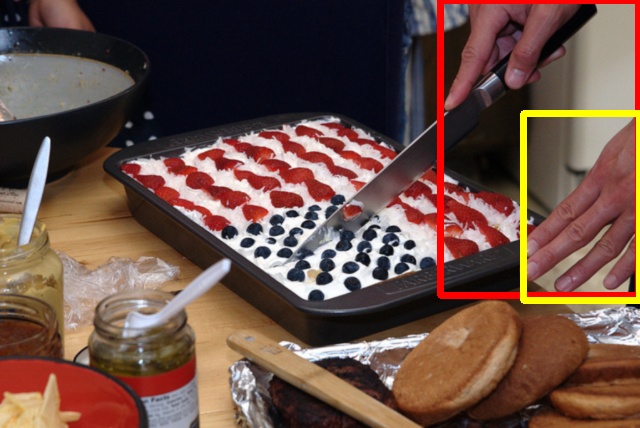} \\
	\end{tabular}
	\caption{\textbf{Failure cases of our method. The red and yellow boxes represent the ground truth and our results, respectively.}}
	\label{fig:failure}
\end{figure*}

In Figure~\ref{fig:final_1}, \ref{fig:final_2}, and \ref{fig:final_3}, we present more qualitative results generated by our full model trained on the RefCOCO, RefCOCO+ and RefCOCOg* datasets.
The proposed method is able to localize objects accurately given the synonymous sentences with paraphrases and also distinguish between sentences that describe different objects.
%
We also provide some failure cases of our method in Figure~\ref{fig:failure}. While the proposed algorithm shows the effectiveness on referring expression comprehension, it still suffers from some unfavorable effects, such as objects sharing similar attributes or ambiguous sentences.

